\pdfoutput=1
\documentclass{article}

     \PassOptionsToPackage{round}{natbib}

\usepackage[final,main]{neurips_2026}

\usepackage[utf8]{inputenc} 
\usepackage[T1]{fontenc}    
\usepackage{url}            
\usepackage{booktabs}       
\usepackage{amsfonts}       
\usepackage{nicefrac}       
\usepackage{microtype}      
\usepackage{xcolor}         

\usepackage{enumerate}
\usepackage{ulem}
\usepackage{bm}
\usepackage{tcolorbox}
\usepackage{enumitem}

\usepackage{amsmath}
\usepackage{amsthm}

\newtheorem{lemma}{Lemma}
\theoremstyle{definition}
\newtheorem{proposition}{Proposition}
\newtheorem{corollary}{Corollary}
\newtheorem{remark}{Remark}

\newcommand{\ESS}{\mathrm{ESS}}

\newcommand{\dd}{\mathrm{d}}
\newcommand{\Dec}{\mathrm{Decode}}

\DeclareMathOperator{\sigmoid}{sigmoid}
\DeclareMathOperator{\std}{std}

\usepackage{algorithm}
\usepackage{algpseudocode}

\title{Boosting Inference with Guided Reasoning: Stochastic\\ Exploration for Recursive Models}

\author{%
  \textbf{Andrew Corbett\textsuperscript{1,2}}\thanks{Correspondence to \texttt{andy.corbett@digilab.ai}.}\quad
  \textbf{Archit Sood\textsuperscript{1}}\quad
  \textbf{Anna Tzatzopoulou\textsuperscript{1}} \\
  \textbf{Sai-Aakash Ramesh\textsuperscript{1,2}}\quad
  \textbf{Tim Dodwell\textsuperscript{1,2}} \\[8pt]
  \textsuperscript{1}digiLab, UK\qquad
  \textsuperscript{2}University of Bristol, UK
}

\begin{document}

\maketitle

\begin{abstract}
  Recent work on recursive architectures has shown that tiny neural networks can be surprisingly powerful on structured reasoning tasks. The trick is to model reasoning trajectories with a latent dynamical system. We argue that the inference-time behaviour of these architectures is best understood as approximate inference over latent reasoning trajectories, with deterministic recursion as the one-particle, zero-noise limit. We make this view operational through \textit{guided reasoning}: stochastic perturbations of the reasoning dynamics propose neighbouring trajectories, and the model's existing early-stopping head reweights them online. The framework yields three diagnostics computable from inference traces: local stability, guide alignment, with a label-free necessary condition, and cloud-token entropy. These predict whether the procedure will help and which of its outputs to trust. On Sudoku-Extreme it lifts exact-solve accuracy from $85.9\%$ to $98.0\%$ without retraining, and on PPBench, a four-type pencil-puzzle benchmark, from $62.5\%$ to $77.9\%$ with no tuning; on Maze-Hard the diagnostics flag a misaligned guide, as validation performance later confirms. The same machinery thus characterises both when recursive reasoning has room to improve at the trajectory level and when the model's internal guide can recover said improvements.
  Code is available at \mbox{\url{https://github.com/digiLab-ai/bigr}}.
\end{abstract}

\section{Introduction}
\label{sec:intro}

While large, compute-intensive models have dominated recent progress on structured reasoning tasks, recursive architectures \citep{wang2025hrm,jolicoeur2025trm}
have shown breakthrough performance, matching large language models, with neural backbones of as little as $7M$ parameters.
This combination of compactness and performance makes them well-positioned for AI deployment in settings with strict latency, privacy, and auditability constraints, where they can offer practical advantages over hyper-scaled alternatives.
We argue that what these architectures do at inference time is best understood as approximate inference over latent reasoning trajectories, with deterministic recursion as the one-particle, zero-noise limit.

Prototypical architectures, such as the Tiny Recursive Model (TRM) of \citet{jolicoeur2025trm}, already contain ingredients that suggest a richer inference procedure at test time, including learned early-stopping $Q$-heads and mixed-frequency latent updates.
These features have emerged predominantly through experimental iteration \citep{wang2025hrm,jolicoeur2025trm}, leaving open how they should be interpreted or used once training is complete.
We study this question through inference-time adaptation: given a frozen pre-trained backbone, how should additional computation be allocated across alternative latent reasoning trajectories, and when can we predict that this allocation will succeed?

We make this view operational without retraining. Stochastic perturbations of the inner recursion induce a proposal over neighbouring latent trajectories; the model's $Q$-head, repurposed as a learned guide, reweights them online via a Feynman--Kac tilt. The deterministic recursion of \citet{jolicoeur2025trm} sits inside this framework as the one-particle, zero-noise limit, allowing direct comparison with the original model. In this work we evaluate our method on structured reasoning tasks. We focus on zero-retraining, meaning the TRM backbone and jointly pre-trained $Q$-head guide are held fixed.

\begin{figure*}[t]
  \centering
    \includegraphics[width=\linewidth]{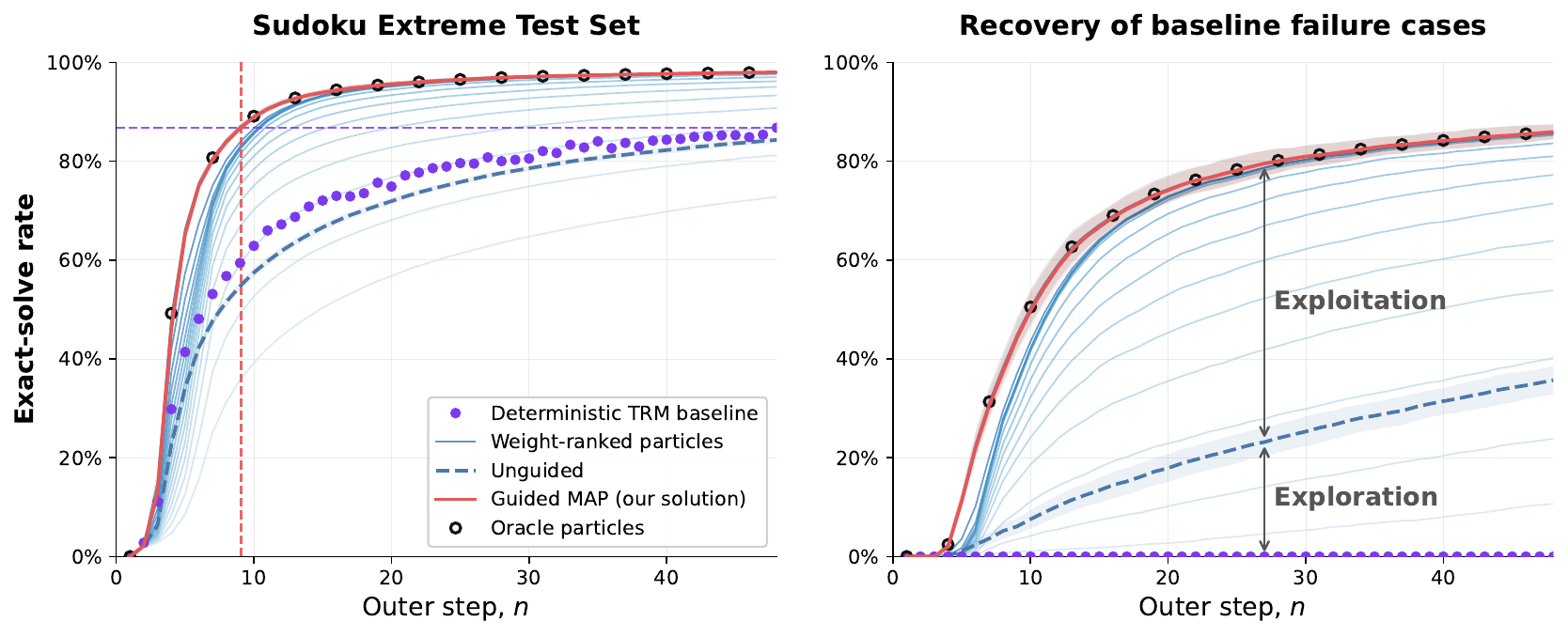}
    \caption{\textbf{Inference-time improvement on Sudoku-Extreme.}
    \textit{Left:} Our method reaches $98.0\!\pm\!0.2\%$ exact-solve accuracy on the test split, improving over the deterministic TRM baseline ($85.9\!\pm\!0.6\%$) and a single unguided rollout by step~$10$.
    \textit{Right:} On deterministic-failure cases, our method solves $85.9\!\pm\!1.5\%$ (with $35.7\!\pm\!2.8\%$ for a single unguided rollout; oracle of the guided cloud $86.0\%$). Results are over $5$ seeds $\times$ $5$ folds; $\pm$ and shaded bands are the fold--seed s.d.}
  \label{fig:main_results}
\end{figure*}

Stochastic exploration substantially increases the number of successful trajectories in the particle cloud, and guided selection identifies them when the guide is aligned.
On Sudoku-Extreme \citep{wang2025hrm}, our method improves exact-solve accuracy from $85.9\!\pm\!0.6\%$ to $98.0\!\pm\!0.2\%$, solving previously out-of-reach failure cases and recovering baseline performance at much lower depths (Figure~\ref{fig:main_results}).

More generally, our procedure helps when two operating conditions are met. Stochastic exploration must be locally stable so alternative trajectories remain recoverable (Proposition~\ref{prop:tube_stability}), and the $Q$-head guide must be aligned with the ability to concentrate particle mass on successful trajectories (Lemma~\ref{lem:bce_aligned}). Alignment admits both a labelled diagnostic, the class-conditional gap $\Delta_n$, and a label-free necessary condition, the in-cloud guide-spread bound \eqref{eq:q_spread_bound}, which upper-bounds the mass shift any tempering can produce. On Sudoku-Extreme both conditions hold, and our method closes nearly the entire oracle gap. On Maze-Hard exploration is stable but the $Q$-head guide is too flat: the spread bound flags this label-free, a priori (Figure~\ref{fig:guide_alignment}), and evaluation confirms the predicted non-improvement (Table~\ref{tab:main_results}). The lesson is that recursive models leave useful trajectory-level headroom on the table, but recovering it requires both stable exploration and a ranking-aligned internal guide.

Our method also yields a third diagnostic: uncertainty statistics for output reliability. We show that Shannon entropies of the final-step and path-integrated token marginals correlate with token error risk and can be used to support selective abstention (\S\ref{sec:token_entropy}, Figure~\ref{fig:token_entropy}).

Our contributions are as follows:

\begin{itemize}
  \item
  \textbf{An approximate inference framework for recursive reasoning models:} we introduce (i) a stochastic proposal kernel for inference-time exploration and (ii) a guided selection algorithm to identify successful trajectories and reweight the distribution; deterministic recursion is recovered exactly in the one-particle, zero-noise limit with the $Q$-head guide.
  \item
  \textbf{Three computable diagnostics to audit performance and reliability:}
  a stability bound (Proposition~\ref{prop:tube_stability}), an alignment characterisation (Lemma~\ref{lem:bce_aligned}), and uncertainty statistics (\S\ref{sec:token_entropy}).
  \item
  \textbf{Training-free improvement in exact-solve accuracy:} guided inference improves exact-solve accuracy from $85.9\%$ to $98.0\%$ on Sudoku-Extreme and from $62.5\%$ to $77.9\%$ on PPBench, without updating model parameters. Compute-matched comparisons isolate the improvements to `guiding' rather than sampling alone, while sequential reweighting improves robustness; see Table \ref{tab:main_results}. On Maze-Hard, our label-free diagnostics correctly identify a guide too flat to exploit the available oracle-particle headroom.
\end{itemize}

\section{Related work}
\label{sec:related}

Both the HRM and TRM algorithms \citep{wang2025hrm,jolicoeur2025trm} build on earlier recurrent relational networks for constraint-satisfaction puzzles \citep{palm2018recurrent}, showing that strong structured reasoning emerges from repeatedly applying a small recurrent core.
\citet{efstathiou2026recursive} characterise TRM's recursion as `searching' in an attractor landscape, observing that on Sudoku-Extreme correct solutions sit within reach of the deterministic path: ``it's in there somewhere''. We add a probabilistic description in which the $Q$-head finds them. On Maze-Hard, where no other line of work reports inference-time improvement, our $Q$-spread diagnostic predicts this behaviour whilst also identifying nearly five points of attainable improvement in the particle cloud: we show ``it's in there, too''. \citet{dillon2026trmspeed} uses halting speed under `Adaptive Computation Time' \citep{graves2016adaptive} as a correctness signal; our method weights the cloud with the head's \textit{value} rather than its \textit{timing}, on fixed weights. These different approaches compose in application.

Most TRM-adjacent work modifies the model or its training, or analyses what the trained model \textit{does}: \citet{mcgovern2025tta_trm} updates TRM weights at test time on each task's demonstration pairs; \citet{asadulaev2025trmpolicy} reinterpret the inner step as policy improvement and modify it; \citet{abubacar2026parameter} freeze a pretrained encoder and train a TRM as a quality estimator; concurrently, \citet{sghaier2026ptrm} add Gaussian noise to the TRM recursion and keep the best of $K$ independent rollouts by the $Q$-head; their strategy is the terminal special case of our filtering approach, with resampling off and the tilt applied once at step $N$. Table~\ref{tab:main_results} compares the two on a matched budget and under a corrupted guide. \citet{royeazar2025trm_arcagi} investigate the inductive biases and test-time compute of the ARC Prize TRM checkpoint. Looped and latent-compute language models vary recursion depth at inference \citep{koishekenov2025encode,zhu2025ouro,bae2025mor}. Our setting is complementary on each of these axes: weights, latent depth, and outputs are all unchanged; we only reallocate compute across stochastic local trajectories and reweight them with the existing $Q$-head.

At a higher level, the work sits alongside inference-time-compute methods for language models that sample chains, branch search, or apply verifiers \citep{wei2022cot,wang2023selfconsistency,yao2023tree,lightman2024letsverify,snell2025reasoning}. We run the same probabilistic machinery (sequential Monte Carlo with a learned twist) but over continuous latent reasoning states inside a small recursive model, rather than over discrete token chains where the analogous twist function appears in twisted SMC \citep{zhao2024twistedsmc}. Recursive computation also relates to Deep Equilibrium Models \citep{bai2019deep} and unrolled policy iteration \citep{behzadian2026unrolled}.

Methodologically, we draw on sequential importance sampling and the Feynman--Kac view of weighted path measures \citep{liu1998sequential,gilks2001following,delmoral2006sequential}; the $Q$-head reweighting is a generalised-Bayes tilt \citep{bissiri2016general}. Our cloud-uncertainty statistics relate to entropy-based uncertainty diagnostics for language models \citep{kuhn2023semantic,farquhar2024detecting}.

\section{Background on recursive models}
\label{sec:setup}

TRM-style architectures \citep{jolicoeur2025trm} collapse the two separate modules of HRM \citep{wang2025hrm} into one module simulating two frequencies. This is applied by a shared backbone $f_\theta$ evaluating two nested loops. An \textit{inner loop} refines a latent state $z\in\mathcal Z$ and an \textit{outer loop} reads out an answer $y\in\mathcal Y$, both in $\mathcal Y=\mathcal Z:=\mathbb R^{L\times D}$ ($L$ sequence length, $D$ feature dimension). We pair them as the joint state $h_n:=(y_n,z_n)\in\mathcal H:=\mathcal Y\times\mathcal Z$ at outer step $n$. In this work we consider the backbone parameters $\theta$ fixed, and $f_\theta$ as a pre-trained deterministic function.
For a fixed embedded input $x \in \mathbb R^{L\times D}$, the inner recursion, for $m=1,\ldots,M-1$, is given by the recursive formula
\begin{equation}
z_{n,1}   = f_{\theta}(x + y_n + z_n),\qquad
z_{n,m+1} = f_{\theta}(x + y_n + z_{n,m}),
\label{eq:inner}
\end{equation}
and the outer step, through \eqref{eq:inner}, reads out the next answer from the final inner latent step $z_{n,M}$:
\begin{equation}
  h_n = (y_n, z_n) \longmapsto \bigl(f_{\theta}(y_n + z_{n,M}),\, z_{n,M}\bigr).
\label{eq:outer}
\end{equation}
Iterating \eqref{eq:inner}--\eqref{eq:outer} from a learned initial state $h_0 = (y_0, z_0)$ produces a deterministic trajectory $h_0, h_1, \ldots, h_N$, with decoded answers $a_n := \Dec(y_n)$ at each outer step $n = 1, \ldots, N$. The inner depth $M$ is fixed at inference by the pre-training value, $M=6$ for Sudoku-Extreme and $M=4$ for Maze-Hard (Appendix~\ref{sec:app_eval_protocol}).

\paragraph{An early-stopping utility.}
The architecture additionally carries a learned scalar $Q$-head $\tilde q_\theta : \mathcal Y \to \mathbb R$, trained under detached BCE supervision as a binary classifier for whether $a_n$ matches the ground-truth solution. \citet{jolicoeur2025trm} use it to trigger Adaptive Computation Time \citep{graves2016adaptive}. We read this as the model's internal logit-scale estimate of halting success, and \S\ref{sec:method} uses it as the natural score on which to tilt the predictive law.

\paragraph{Notation.}
We identify tensors in $\mathbb R^{L\times D}$ with their vectorisations in $\mathbb R^{LD}$ and write $\|\cdot\|$ for the induced Euclidean / Frobenius norm; the same symbol is used throughout \S\ref{sec:method}--\S\ref{sec:theory}. For isotropic Gaussian noise $\xi_m \sim \mathcal N(0, I_{LD})$ the second-moment bound in Proposition~\ref{prop:tube_stability} satisfies $\nu^2 = \mathbb E\|\xi_m\|^2 = LD$. We write $\sigmoid(y) := (1+e^{-y})^{-1}$ for the sigmoid link function.

\section{Guided reasoning for recursive models}
\label{sec:method}

We build on the deterministic recursion of \S\ref{sec:setup} to construct a path measure over latent reasoning trajectories: a stochastic kernel, generalising the inner-recursion \eqref{eq:inner}, proposes neighbouring trajectories (\S\ref{sec:inner}); the $Q$-head reweighting tilts the predictive law toward correct solutions (\S\ref{sec:outer}); and a bootstrap particle approximation makes the resulting Feynman--Kac flow tractable (\S\ref{sec:particle_impl}).

\subsection{Exploration: the stochastic proposal}
\label{sec:inner}

To widen the inner TRM recursion for exploration, let $\xi_{n,0:M}$ be independent mean-zero noise variables and $\sigma\ge 0$ the exploration scale. Define
\begin{equation}
\zeta_{n,1} = z_{n,1} + \sigma\xi_{n,0},\qquad
\zeta_{n,m+1} = f_{\theta}(x+y_n+\zeta_{n,m}) + \sigma\xi_{n,m},
\label{eq:stoch_inner}
\end{equation}
for $m=1,\ldots,M-1$, and the stochastic outer update
\begin{equation}
h_n = (y_n, z_n)
\longmapsto
\bigl(f_{\theta}(y_n+\zeta_{n,M}) + \sigma\xi_{n,M},\,\zeta_{n,M}\bigr).
\label{eq:stoch_outer_update}
\end{equation}
The proposal kernel is then the distribution induced by this random transition:
\begin{equation}
K_{f_{\theta},x}(A\mid h_{n})
:=
\mathbb{P}\bigl(\bigl(f_{\theta}(y_n+\zeta_{n,M}) + \sigma\xi_{n,M},\,\zeta_{n,M}\bigr)\in A\bigr),
\label{eq:transition_kernel}
\end{equation}
for measurable $A\subseteq\mathcal H$, with implicit dependence on the inner depth $M$ and noise scale $\sigma$ through \eqref{eq:stoch_inner}.
At $\sigma=0$ the kernel collapses to the Dirac measure on \eqref{eq:outer}, recovering the deterministic TRM update. For $\sigma>0$ it widens the inner trajectory locally without changing backbone weights.

\subsection{Exploitation: the $Q$-head tilt}
\label{sec:outer}

Local exploration alone defines the proposal kernel \eqref{eq:transition_kernel}; we select among proposals by reweighting with the $Q$-head $\tilde q_\theta$, instantiated as a Feynman--Kac tilt \citep{delmoral2006sequential} of the predictive law.
Let $\pi_n$ denote the law of the guided state after $n$ outer steps. Propagation through the proposal kernel gives the predictive law
\begin{equation}
\pi_{n+1\mid n}(\dd h')
=
\int_{\mathcal H} K_{f_{\theta},x}(\dd h' \mid h)\,\pi_n(\dd h),
\label{eq:predictive_law}
\end{equation}
which we tilt by the $Q$-head. Working on the log-probability scale,
\begin{equation}
q_\theta(h):=\log\sigmoid\bigl(\tilde q_\theta(y)\bigr),
\label{eq:guide_score}
\end{equation}
where $h=(y,z)$, so that $G_\beta(h):=\exp(\beta q_\theta(h))\in(0,1]$ is a proper potential for every inverse temperature $\beta\ge 0$. The guided update is the exponentially tilted $\beta$-law
\begin{equation}
\pi_{n+1,\beta}(\dd h)
\propto
G_\beta(h)\,\pi_{n+1\mid n}(\dd h),
\label{eq:guided_update}
\end{equation}
a single mutation--selection step of the corresponding Feynman--Kac flow. Applying \eqref{eq:guided_update} once, at step $N$ only and without resampling, reduces to a best-in-cloud selection by the $Q$-head; see Appendix~\ref{app:selection_rules}.

\subsection{Finite-particle implementation}
\label{sec:particle_impl}

We approximate \eqref{eq:predictive_law} with an empirical measure $\pi_n(\dd h)\approx\sum_{s=1}^{S} w_n^{(s)}\delta_{h_n^{(s)}}(\dd h)$. Propagating each particle through the proposal kernel gives the predictive mixture
\begin{equation}
\pi_{n+1\mid n}(\dd h)
\approx
\sum_{s=1}^{S} w_n^{(s)}\,K_{f_{\theta},x}(\dd h \mid h_n^{(s)}),
\label{eq:moe}
\end{equation}
and the update therefore decomposes into exploration (\S\ref{sec:inner}) and guide-based exploitation \eqref{eq:guided_update},
\begin{align}
\text{Exploration: }&\quad
h_{n+1}^{(s)} \sim K_{f_{\theta},x}(\cdot \mid h_n^{(s)}),
\label{eq:explore}\\
\text{Exploitation: }&\quad
w_{n+1}^{(s)} \propto w_n^{(s)}\,G_\beta\bigl(h_{n+1}^{(s)}\bigr).
\label{eq:exploit}
\end{align}

\begin{algorithm}[h]
\caption{\textbf{Guided reasoning for recursive models.} Hyperparameters: $S$ particles, outer depth $N$, inner depth $M$, noise scale $\sigma\geq 0$, inverse temperature $\beta\geq 0$, ESS threshold $\tau_{\ESS}\in(0,1]$.}
\label{alg:guided_inference}
\small
\begin{algorithmic}[1]
\Require input $x$, initial state $h_0$, backbone $f_\theta$, guide $q_\theta$, decoder `$\Dec$'
\Function{TrajectorySampler}{$ h = ( y,  z)$}
\Comment{sample the proposal kernel \eqref{eq:transition_kernel}}
  \State draw $\xi_{0:M}\sim\mathcal N(0, I_{LD})$
  \State $\zeta_1 \gets f_\theta(x +  y +  z) + \sigma\xi_0$
  \For{$m=1,\ldots,M-1$}
    \State $\zeta_{m+1} \gets f_\theta(x +  y + \zeta_m) + \sigma\xi_m$ \Comment{stochastic inner-recursion \eqref{eq:stoch_inner}}
  \EndFor
  \State \Return $\bigl(f_\theta( y + \zeta_M) + \sigma\xi_M,\; \zeta_M\bigr)$ \Comment{stochastic read-out step \eqref{eq:stoch_outer_update}}
\EndFunction
\Statex
\Statex \textbf{Outer loop: $q_\theta$-tempered particle filter} \Comment{tilted Feynman--Kac flow \eqref{eq:guided_update}}
\State Initialise $ h_0^{(s)} = h_0$, $ w_0^{(s)} = 1/S$ for $s=1,\ldots,S$ \Comment{equally weighted initialisation}
\For{$n=0,\ldots,N-1$}
  \For{$s=1,\ldots,S$}
    \State $h_{n+1}^{(s)} \gets \Call{TrajectorySampler}{ h_n^{(s)}}$
    \State $\widetilde w_{n+1}^{(s)} \gets  w_n^{(s)} \exp\!\bigl(\beta\,q_\theta(h_{n+1}^{(s)})\bigr)$ \Comment{weight-guiding step \eqref{eq:guided_update}}
  \EndFor
  \State $w_{n+1}^{(s)} \gets \widetilde w_{n+1}^{(s)} \big/ \sum_r \widetilde w_{n+1}^{(r)}$ $\forall s=1,\ldots,S$ \Comment{normalise across cloud}
  \If{$\mathrm{ESS}_{n+1}/S < \tau_{\ESS}$}
      \State $h_{n+1}^{(s)} \gets h_{n+1}^{(a_s)}$, $w_{n+1}^{(s)} \gets 1/S$
      \Comment{systematic resampling; ancestors $a_s$}
  \EndIf
\EndFor
\State \Return {$\widehat a_N(x) := \Dec\bigl(y_N^{(s^\star)}\bigr)$, $s^\star \in \arg\max_s w_N^{(s)}$} \Comment{highest-weight particle}
\end{algorithmic}
\end{algorithm}

Algorithm~\ref{alg:guided_inference} gives the bootstrap particle approximation \eqref{eq:moe}.
\textsc{TrajectorySampler} stochastically generalises the inner TRM recursion \eqref{eq:inner}--\eqref{eq:outer}; the outer $Q$-tempered particle filter generalises single-path rollout to a weighted cloud, with deterministic TRM as the zero-noise, single-particle path.
Resampling triggers only when the effective sample size
\begin{equation}
\mathrm{ESS}_n := \bigl(\sum_s (w_n^{(s)})^2\bigr)^{-1}
\end{equation}
falls below $\tau_{\ESS}S$, using the systematic protocol \citep{kitagawa1996monte,carpenter1999improved}: ancestor indices $a_{1:S}$ are drawn from the cumulative-weight CDF at stratified levels with a single shared uniform offset, replicating high-weight particles and dropping low-weight ones at the uniform new weight $1/S$. The terminal decoder returns the answer of the highest-weight particle, a particle-level MAP estimate.

\begin{remark}[Cost and the deterministic limit]
\label{rem:deterministic_limit}
The per-particle steps in Algorithm~\ref{alg:guided_inference} are independent across $s$ and can batch along the sample axis. Measured on one H100 GPU at $N=48$, a puzzle costs $0.014$\,s at $S=1$ and $0.058$\,s at $S=16$, sixteen times the forward calls for roughly four times the wall-clock time, with peak memory growing linearly from $84$ to $554$\,MB (Table~\ref{tab:cost}). Stopping early when the particle cloud's largest guide confidence exceeds $0.95$ matches full-depth accuracy at $14.5\%$ of the $N=48$ budget on Sudoku-Extreme Test and $35\%$ on DF (Appendix~\ref{app:depth_cost}). With $S=1$ and $\sigma=0$, the algorithm reduces exactly to the deterministic TRM \citep{jolicoeur2025trm}.
\end{remark}

\section{Operating theory and diagnostics}
\label{sec:theory}

The probabilistic formulation in \S\ref{sec:method} lets us characterise when the $Q$-head's re-ranking is aligned with successful trajectories. The theory yields two conditions for whether guided exploration helps, \textit{tube stability} (\S\ref{sec:local_stability}) and \textit{guide alignment} (\S\ref{sec:bce_alignment}), plus a third, output-side diagnostic, \textit{token-marginal entropy} (\S\ref{sec:token_entropy}), on the resulting particle cloud.

\subsection{Local tube stability of stochastic trajectories}
\label{sec:local_stability}

\begin{figure*}[h]
  \centering
  \includegraphics[width=\textwidth]{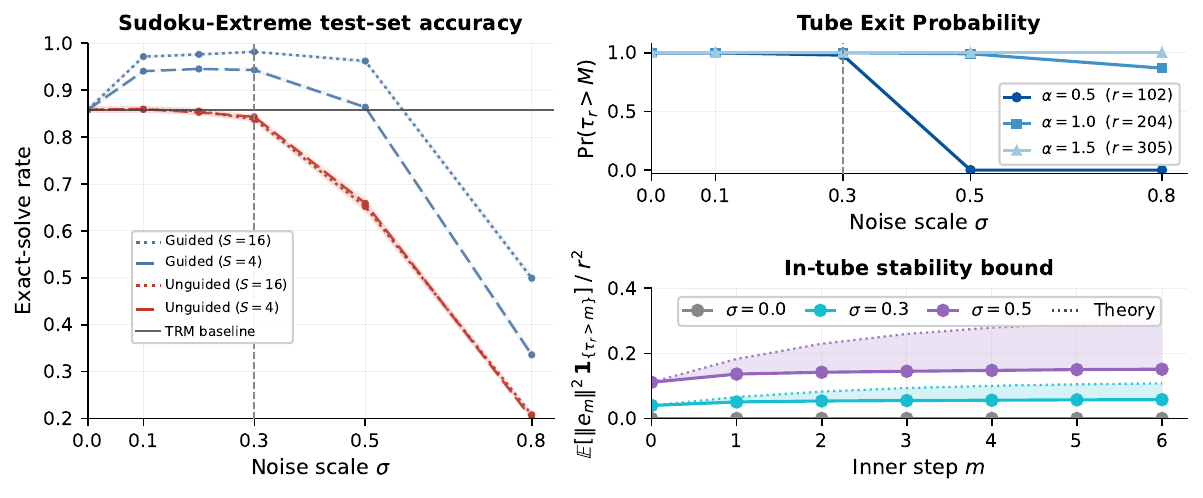}
  \caption{\textbf{Local stability and the noise sweet spot on Sudoku-Extreme.}
  \textit{Left}: terminal exact-solve rate vs.\ $\sigma$ for unguided ($\beta=0$) and guided ($\beta=0.25$) inference at $S\in\{4,16\}$.
  \textit{Right, top}: tube-survival probability $\Pr(\tau_r>M)$ vs.\ noise scale for radii $r=\alpha\sqrt{LD}$.
  \textit{Right, bottom}: average in-tube deviations $\mathbb{E}[\|e_m\|^2\mathbf{1}_{\{\tau_r>m\}}]/r^2$ per inner step at $\alpha=1.5$; dotted curves give the bound \eqref{eq:tube_pre_exit_bound} at $\hat\rho_r$, the empirical local Lipschitz constant on the $r$-tube estimated as the maximum secant ratio over $16$ random unit-direction probes at each point $z_m$, $0\le m\le M$.
  At $\sigma=0.3$, accuracy is maximised and tube-containment remains high, with deviations within the theory bound.}
\label{fig:inner_deviation}
\end{figure*}

We analyse the local properties of the inner recursion \eqref{eq:stoch_inner}. The resulting predictions are measured in Figure~\ref{fig:inner_deviation}.
Fix $x\in\mathbb{R}^{L\times D}$ and $h_n\in\mathcal{H}$, and suppress these, as well as subscripts $n$, from the notation. Let $z_{1:M}$ denote the deterministic inner trajectory and $\zeta_{1:M}$ the stochastic rollout of \eqref{eq:stoch_inner}. Write $e_m := \zeta_m-z_m$ and
\begin{equation}
\tau_r := \inf\{m\in\{1,\dots,M\}:\|e_m\|>r\}
\label{eq:em_taur}
\end{equation}
for the deviation at step $m$ and the first-exit step from the $r$-tube around the deterministic path, respectively, with the convention $\tau_r=M+1$ when the set is empty.

\begin{proposition}[Tube stability]
\label{prop:tube_stability}
Assume there exist $r>0$ and $\rho_r\in[0,1)$ such that $z\mapsto f_{\theta}(x+y_n+z)$ is $\rho_r$-Lipschitz on the $r$-tube $\bigcup_{m=1}^{M}\{z:\|z-z_m\|\le r\}$, and that the noise variables are independent, mean-zero, with $\mathbb{E}\|\xi_m\|^2\le\nu^2$. Then for $m=1,\ldots,M$ the average in-tube deviations conform to the geometric decay bound
\begin{equation}
\mathbb{E}\bigl[\|e_m\|^2\mathbf 1_{\{\tau_r>m\}}\bigr]
\;\le\;
\sigma^2\nu^2\sum_{j=0}^{m-1}\rho_r^{2j}
\;\le\;
\frac{\sigma^2\nu^2}{1-\rho_r^2}.
\label{eq:tube_pre_exit_bound}
\end{equation}
\end{proposition}

\begin{corollary}[Finite-horizon tube-exit probability]
  \label{cor:tube_exit}
Under the hypotheses of Proposition~\ref{prop:tube_stability} we have
\begin{equation}
\mathbb{P}(\tau_r\le M)
\;\le\;
\frac{M}{r^2}\frac{\sigma^2\nu^2}{1-\rho_r^2}.
\label{eq:tube_exit_bound}
\end{equation}
\end{corollary}

Proposition~\ref{prop:tube_stability} controls the deviation $e_m$ while the rollout stays in the $r$-tube; Corollary~\ref{cor:tube_exit} then quantifies how often that control applies. Together \eqref{eq:tube_pre_exit_bound}--\eqref{eq:tube_exit_bound} identify a regime in which stochastic widening remains controlled. Fixed $M$, larger $r$, smaller $\sigma$, and stronger local contraction all tighten both bounds. The in-tube bound does not itself depend on $M$, and both bounds need only independent, mean-zero noise of finite second moment (Remarks~\ref{rem:noise_assumptions} and~\ref{rem:uniform_in_M} in Appendix~\ref{app:proofs}).
We evaluate these bounds empirically in Figure~\ref{fig:inner_deviation}, showing that the deviation bound is not exceeded and that the probability of remaining within the tube is high at moderate noise levels.


\subsection{Alignment of the $Q$-head guide}
\label{sec:bce_alignment}

Here we show that the BCE-trained $Q$-head is aligned with its training label in the integrated-covariance sense of~\eqref{eq:gibbs_covariance} below.
The family $\{\pi_{n+1,t}\}_{t\in[0,\beta]}$ in \eqref{eq:guided_update} is a one-parameter exponential tilt, such that bounded $\varphi:\mathcal H\to\mathbb R$ satisfies the tilt-derivative identity \citep{gelman1998simulating}
\begin{equation}
\frac{\mathrm d}{\mathrm d t}\mathbb E_{\pi_{n+1,t}}[\varphi(h)]
=
\operatorname{Cov}_{\pi_{n+1,t}}\!\bigl(\varphi(h),\,q_\theta(h)\bigr).
\label{eq:gibbs_covariance}
\end{equation}
We call a guide $q_\theta$ \textit{aligned} with $\varphi$ on $[0,\beta]$ if and only if $\operatorname{Cov}_{\pi_{n+1,t}}(\varphi,q_\theta)\geq 0$ throughout. Equivalently, $\mathbb E_{\pi_{n+1,t}}[\varphi(h)]$ is non-decreasing in $t$ on the tilting path $[0,\beta]$. Integrating yields an improvement measure against an `unguided' baseline. Appendix~\ref{app:guide_quality} measures the calibration and the discrimination of the guide by outer step on each benchmark.

\begin{lemma}[BCE alignment]
\label{lem:bce_aligned}
Let $Y\in\{0,1\}$ and let $0<\eta<1$ be a random variable with $\mathbb E[Y\mid\eta]=\eta$. For any strictly increasing $\psi$, the guide $q_\theta=\psi(\eta)$ satisfies $\operatorname{Cov}_{\pi_{n+1,t}}(Y,\,q_\theta)\ge 0$ for all $t\in[0,\beta]$. In particular, for $q_\theta=\log\eta$, the population BCE optimum among guides on which $\eta$ is the conditional expectation of $Y$, is aligned with $Y$.
\end{lemma}

\begin{figure*}[h]
  \centering
  \includegraphics[width=\textwidth]{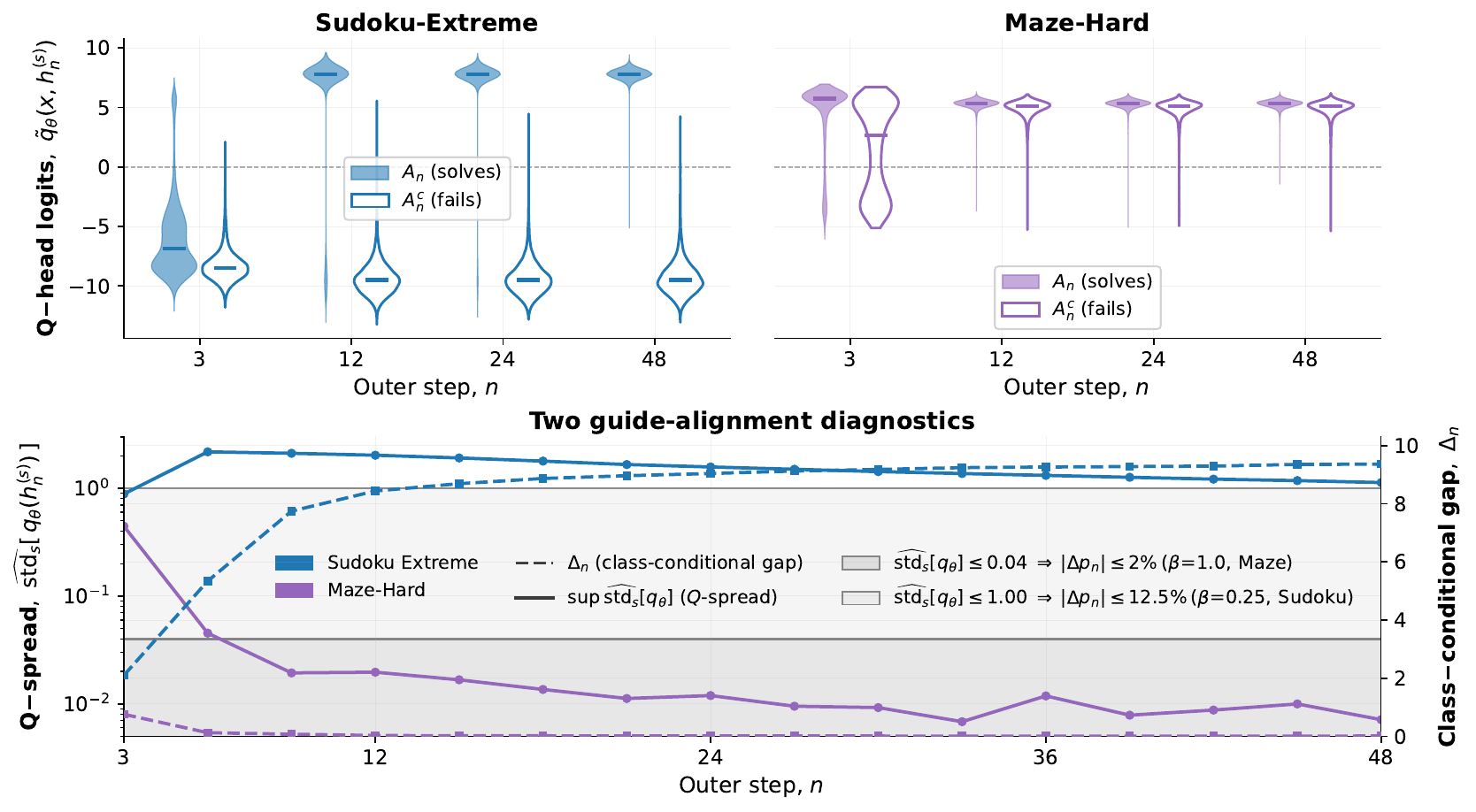}
  \caption{\textbf{Guide-alignment diagnostics.}
  \label{fig:guide_alignment}
  \textit{Top}: the raw $\tilde q_\theta$ logits split by puzzle-solve outcome over outer step $n$; strong class separation on Sudoku-Extreme (\textit{left}), failure to discriminate on Maze-Hard (\textit{right}).
  \textit{Bottom}: the post-hoc class-conditional gap $\hat\Delta_n$ \eqref{eq:delta} (right-axis, dotted) and the $Q$-spread bound \eqref{eq:q_spread_bound} (left-axis, solid) for the mass shift $|p_n(\beta)-p_n(0)|$, both in the guide-score units $q_\theta=\log\sigmoid(\tilde q_\theta)$ on which the tilt acts. On Maze-Hard the bound is prohibitive: $\le 2\%$ at $\beta=1$, tightening to $\le 0.5\%$ at $\beta=0.25$; on Sudoku-Extreme the bound exceeds $12.5\%$ at $\beta=0.25$, in line with the observed improvement (Figure~\ref{fig:main_results}).
  }
\end{figure*}

TRM's $Q$-head is trained under BCE, so Lemma~\ref{lem:bce_aligned} applies with $\eta=\sigmoid(\tilde q_\theta(y))$ whenever the head attains the population BCE optimum under \eqref{eq:guided_update}. Note that training does not guarantee this: the optimum is a population statement over the training distribution of puzzles, whereas the covariance in \eqref{eq:gibbs_covariance} is taken over the particle cloud of a single test instance. Lemma~\ref{lem:bce_aligned} therefore describes the guide at the BCE optimum rather than the trained head itself (cf.\ Appendix~\ref{sec:app_proof_of_lemma}), so we test alignment from inference traces, both with labels (post-hoc) and label-free; Maze-Hard (\S\ref{sec:experiments}) is a case where the idealisation fails. For $H\sim\pi_{n+1,t}$, write
$p_n(t):=\mathbb P_{\pi_{n+1,t}}(H\in A_n)$
for the probability mass on the success set $A_n\subseteq\mathcal H$ under the tilted law.
We then introduce two scalar quantities derived from~\eqref{eq:gibbs_covariance} that test alignment at outer step~$n$ (Figure~\ref{fig:guide_alignment}).

\paragraph{Class-conditional gap.}
Here we define a post-hoc diagnostic to measure \eqref{eq:gibbs_covariance} in the binary-outcome case; see Appendix~\ref{sec:app_gap} for derivation and discussion.
For $\varphi=\mathbf 1_{A_n}$, the covariance~\eqref{eq:gibbs_covariance} factorises as $p_n(t)\bigl(1-p_n(t)\bigr)\,\Delta_n(t)$ with
\begin{equation}
\label{eq:delta}
\Delta_n(t):=\mathbb E_{\pi_{n+1,t}}[q_\theta\mid A_n]-\mathbb E_{\pi_{n+1,t}}[q_\theta\mid A_n^c].
\end{equation}
Alignment on $[0,\beta]$ is therefore equivalent to the scalar condition $\Delta_n(t)\ge 0$ for all $t\in[0,\beta]$.
The sample diagnostic $\hat\Delta_n$ requires success labels and is therefore a post-hoc score.
Figure~\ref{fig:guide_alignment} tracks~\eqref{eq:delta} on a particle cloud and shows the predicted mass shift emerging as soon as the $Q$-head separates successful from unsuccessful trajectories.
Intuitively, $\hat\Delta_n$ is the gap between the mean guide score $q_\theta$ on solving particles and on failing ones. It accounts for the `space' the guide has to discriminate.

\paragraph{Guide spread over the cloud.}
A label-free necessary condition for a non-trivial tempering improvement can be observed in the cloud directly. Applying Cauchy--Schwarz to~\eqref{eq:gibbs_covariance} with $\std(\mathbf 1_{A_n})\le\tfrac12$ yields the integrated bound
\begin{equation}
  \label{eq:q_spread_bound}
\bigl|\,p_n(\beta)-p_n(0)\,\bigr|\;\le\;\tfrac{\beta}{2}\sup_{t\in[0,\beta]}
\widehat{\std}_s\!\bigl[q_\theta(h_{n+1}^{(s)})\bigr],
\end{equation}
where $\widehat{\std}_s$ denotes the empirical standard deviation over the particle approximation of $\pi_{n+1,t}$, standing in for the population quantity $\std_{\pi_{n+1,t}}[q_\theta]$ of the exact bound, and the supremum runs over the tilting path $t\in[0,\beta]$.
A non-negligible $\widehat{\std}_s$ is therefore necessary (but not sufficient) for a meaningful improvement at outer step~$n$. The spread bounds the rate of mass shift $|p_n(\beta)-p_n(0)|$ in either direction, so a small spread can rule out gains a priori.

In Figure~\ref{fig:guide_alignment} (bottom), the Maze-Hard $Q$-spread satisfies $\sup_{t\leq\beta}\widehat{\std}_s[q_\theta]\le 0.04$. By \eqref{eq:q_spread_bound}, this caps the achievable mass shift $|\Delta p_n|$ at $2\%$ for $\beta=1$ and $0.5\%$ for $\beta=0.25$, well below the $5\%$ oracle headroom on Maze-Hard. On Sudoku-Extreme, by contrast, the spread is large enough that \eqref{eq:q_spread_bound} permits at least a $12.5\%$ mass shift at $\beta=0.25$, consistent with the observed improvement (Figure~\ref{fig:main_results}).

\subsection{Token entropy as an uncertainty diagnostic}
\label{sec:token_entropy}

\begin{figure*}[h]
\centering
\includegraphics[width=0.9\textwidth]{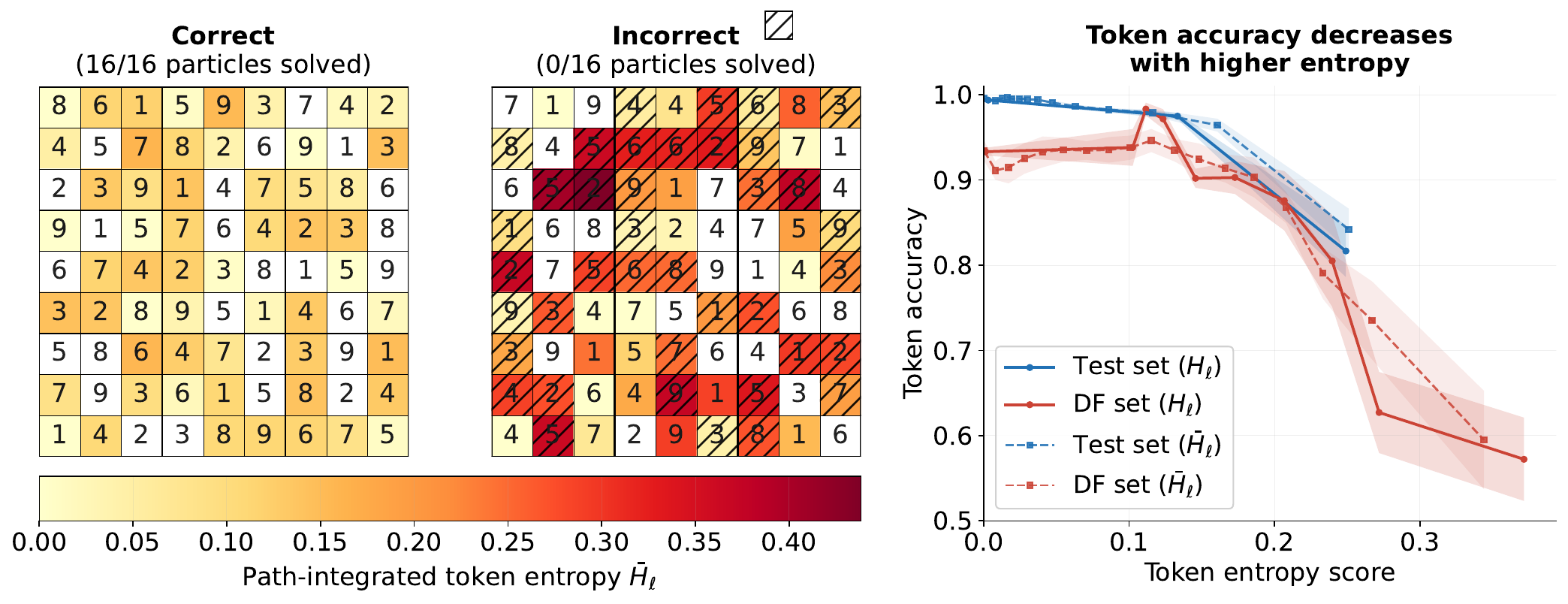}
\caption{\textbf{Token entropy as a risk-ranking diagnostic on Sudoku-Extreme.}
\textit{Left, middle}: solved and unsolved test puzzles coloured by per-cell path-integrated entropy $\bar H_\ell$ (shaded: incorrect tokens; white: given cells).
\textit{Right}: binned token accuracy vs.\ uncertainty on Test and DF for $H_\ell$ and $\bar H_\ell$; bands are $95\%$ CIs.
Token-level AUROC for $H_\ell$: Test $0.834\pm 0.011$, DF $0.710\pm 0.014$.}
\label{fig:token_entropy}
\end{figure*}

Modelling a particle distribution \eqref{eq:predictive_law} also yields in-cloud uncertainty diagnostics at the token level. Let $\mathrm{Dec}:\mathcal H\to\{1,\dots,C\}^L$ be the deterministic decoder. For each token position $\ell$, the cloud induces a class marginal $p_\ell(\beta;n):=\bigl(p_{\ell,1}(\beta;n),\ldots,p_{\ell,C}(\beta;n)\bigr)$ with $p_{\ell,k}(\beta;n):=\mathbb P_{\pi_{n+1,\beta}}(\mathrm{Dec}(h)_\ell=k)$, read off the particle weights at no extra cost. We summarise its dispersion by the normalised Shannon entropy, both at the terminal step and aggregated over outer steps $n_1,\ldots,n_J$:
\begin{align}
\text{Terminal: } & \quad
H_\ell
:=
\frac{1}{\log C}\, H\bigl(p_\ell(\beta;N)\bigr),
\label{eq:token_entropy_terminal}
\\\text{Aggregated: } & \quad
\bar H_\ell
:=
\frac{1}{J\log C}\sum_{j=1}^{J} H\bigl(p_\ell(\beta;n_j)\bigr).
\label{eq:token_entropy_aggregated}
\end{align}
We treat $H_\ell$ and $\bar H_\ell$ as risk-ranking diagnostics over token-level errors; empirical AUROCs and selective-prediction curves are in Figure~\ref{fig:token_entropy}.

\section{Experimental evaluation}
\label{sec:experiments}

We evaluate our guided reasoning method, as described in Algorithm~\ref{alg:guided_inference}, on two fixed TRM models trained on Sudoku-Extreme and Maze-Hard, respectively, and on a third that we train for PPBench, a pencil-puzzle benchmark \citep{waugh2026ppbench}. Particle count, horizon and resampling threshold are validation-tuned on Sudoku-Extreme and transferred to Maze-Hard; the Maze-Hard noise scale and inverse temperature are fixed a priori without Maze-Hard data, with the diagnostics of \S\ref{sec:theory} acting as a verifier; reported numbers are single-shot test-set evaluations. See Appendix~\ref{sec:app_eval_protocol} for further details.

The diagnostics of \S\ref{sec:theory} predict, before any evaluation, where the procedure should help. The inner-deviation bound \eqref{eq:tube_pre_exit_bound}--\eqref{eq:tube_exit_bound} is conformed to on both benchmarks at moderate noise (Figure~\ref{fig:inner_deviation}). The $Q$-spread bound \eqref{eq:q_spread_bound} flags the Maze-Hard $Q$-head as too flat to produce a non-trivial tempering improvement, while Sudoku-Extreme clears both the $Q$-spread and class-conditional thresholds (Figure~\ref{fig:guide_alignment}). The diagnostics therefore only predict non-improvement on Maze-Hard.

Table~\ref{tab:main_results} confirms both predictions. On Sudoku-Extreme, guided MAP closes nearly the entire oracle gap ($85.9\%\to 98.0\%$), recovering $85.9\%$ on the deterministic-failure split where the baseline scores zero. On Maze-Hard, stochastic exploration exposes nearly five points of oracle-particle headroom ($91.3\%$ vs.\ the deterministic $86.6\%$), but the misaligned $Q$-head does not concentrate mass on successful trajectories, so guided MAP does not improve on a single rollout ($85.6\%$ against $85.2\%$). Token-marginal entropy ranks misclassification risk on Sudoku-Extreme (Figure~\ref{fig:token_entropy}): abstaining on the upper-decile $H_\ell$ raises retained exact-solve rate to $100.0\!\pm\!0.0\%$ on Test and $99.7\!\pm\!0.3\%$ on DF.

\begin{table*}[h]
  \centering
  \small
  \caption{\textbf{Terminal exact-solve rates at matched budgets.} Below, $S=16$, $N=48$. For every stochastic evaluation there are $5{,}376$ backbone calls per Sudoku-Extreme puzzle against $336$ for the deterministic TRM. Core runs are evaluated on a $5$ seeds $\times$ $5$ folds design (the deterministic TRM varies over folds only). Particle cloud statistics re-read the stored unguided ($\beta=0$) particles generated on the full splits ($5$ folds of one seed). Puzzle counts for every evaluation are in Table~\ref{tab:puzzle_counts}; the rules, and paired same-run differences, in Appendix~\ref{app:selection_rules}. PPBench evaluates on $5$ seeds on the $253$-puzzle validation split (Appendix~\ref{app:ppbench}). Last column: the DF clouds rescored after i.i.d.\ Gaussian noise of s.d.\ $16$ on every raw guide logit (Appendix~\ref{app:selection_rules}). Values are $\%$ solve rates.}

  \label{tab:main_results}
  \setlength{\tabcolsep}{4pt}
  \begin{tabular}{@{}lccccc@{}}
    \toprule
    Method & Sudoku Test & Sudoku DF & Maze Test & PPBench val & DF, corrupted guide \\
    \midrule
    \multicolumn{6}{@{}l}{\textit{Core runs}} \\
    Deterministic TRM & $85.9\!\pm\!0.6$ & $\phantom{0}0.0$ & $86.6\!\pm\!2.1$ & $62.5$ & -- \\
    Single rollout ($\beta=0$) & $84.3\!\pm\!0.7$ & $35.7\!\pm\!2.8$ & $85.2\!\pm\!2.5$ & $63.6\!\pm\!1.6$ & -- \\
    Sequential guiding (ours) & $98.0\!\pm\!0.2$ & $85.9\!\pm\!1.5$ & $85.6\!\pm\!2.8$ & $77.9\!\pm\!0.6$ & -- \\
    \midrule
    \multicolumn{6}{@{}l}{\textit{Particle cloud statistics}} \\
    Plurality vote & $96.8\!\pm\!0.7$ & $79.5\!\pm\!1.2$ & $86.0\!\pm\!4.1$ & $67.7\!\pm\!1.4$ & -- \\
    Terminal-$Q$ (PTRM's rule) & $98.1\!\pm\!0.5$ & $88.4\!\pm\!0.6$ & $86.9\!\pm\!4.0$ & $79.5\!\pm\!0.8$ & $65.3\!\pm\!0.9$ \\
    Path-sum tilt (no resampling) & $98.1\!\pm\!0.5$ & $88.1\!\pm\!0.9$ & $85.7\!\pm\!3.8$ & $78.7\!\pm\!0.7$ & $83.9\!\pm\!1.6$ \\
    Oracle (any correct particle) & $98.1\!\pm\!0.5$ & $88.4\!\pm\!0.6$ & $91.3\!\pm\!2.7$ & $82.8\!\pm\!0.7$ & -- \\
    \bottomrule
  \end{tabular}
\end{table*}

\paragraph{Selection rules at matched compute.}
The stochastic rows of Table~\ref{tab:main_results} spend the same backbone calls and differ only in how selection is performed in the particle cloud. Three findings follow. First, the gain is the guide, not just the sampling. On PPBench \citep{waugh2026ppbench}, where we train a TRM on four puzzle types and transfer the Sudoku-Extreme configuration without tuning (Appendix~\ref{app:ppbench}), the plurality vote recovers $4$ of the $19$ points between a single rollout and the oracle, and each guide-consistent selection rule recovers $14$ to $16$, on every type. Second, with a near-perfect guide the guided rules tie. On Sudoku-Extreme the terminal guide score separates correct from incorrect particles with AUROC $1$ (Appendix~\ref{app:guide_quality}), and the best-of-$K$ selection rule of PTRM \citep{sghaier2026ptrm}, the path-sum tilt and sequential guiding agree to within $0.2$ points on the Test split (Appendix~\ref{app:selection_rules}). Third, the selection rules part when the guide degrades. With Gaussian noise of s.d.\ $16$ on every raw logit, the path-sum tilt holds $83.9\%$ on DF while terminal selection falls to $65.3\%$: the sum over $48$ steps averages noise that one terminal read-out cannot. Depth is no substitute for sampling (Appendix~\ref{app:depth_cost}).

\paragraph{Limitations and scope.}
Empirical evaluation covers Sudoku-Extreme, Maze-Hard and PPBench, all structured puzzles solved by a backbone with a trained halting head, which is the regime the guide requires; PPBench needed such a backbone trained first. Language tasks are not evaluated, and ARC-AGI \citep{arcprize2025arc2}, a common LLM and TRM-comparison benchmark, remains a clear next evaluation step despite requiring a far higher compute budget.
The framework is inference-only over a fixed backbone with fixed outer horizon $N$; coupling with continued training, model surgery, or non-recursive backbones is left to future work, and weight-updating test-time adaptation \citep{mcgovern2025tta_trm} is complementary rather than compared.
Our theory assumes a local-Lipschitz condition of $f_\theta$ on the relevant $r$-tube (Proposition~\ref{prop:tube_stability}) and a $Q$-head aligned with success in the sense of Lemma~\ref{lem:bce_aligned}, which BCE training motivates but does not guarantee; the Maze-Hard misalignment shows when these conditions bind, and transfer to other recursive architectures remains open.
The label-free spread bound \eqref{eq:q_spread_bound} is necessary but not sufficient: a guide that clears the threshold could still fail in ways spread alone cannot detect.
Our $Q$-head was trained for early stopping rather than trajectory ranking; a purpose-trained guide is the natural follow-up (\S\ref{sec:discussion}).

\section{Discussion and broader impact}
\label{sec:discussion}

The original motivation for this work was to explore what an approximate Bayesian interpretation of recursive reasoning \citep{jolicoeur2025trm} might reveal about the elusive mechanics of the architecture. The investigation produced a Feynman--Kac-inspired theory that elevates the $Q$-head from afterthought to central guide for the selection of latent reasoning trajectories. The same machinery characterises both when recursive reasoning has trajectory-level headroom and when the model's internal guide can recover it, a thesis we believe extends to recursive reasoning more broadly.

What surprised us was the immediate success of both ingredients on Sudoku-Extreme: modest stochastic exploration produced successful trajectories, and the $Q$-head correctly weighted particles towards these (Figure~\ref{fig:main_results}). Maze-Hard was a step into the unknown as no prior work reports inference-time improvements there \citep[cf.][]{mcgovern2025tta_trm}. Stochastic exploration again exposed oracle-level headroom, but, as the theory predicted, the guide could not identify it.
The difference in character between these two benchmarks comes down to the guide. Improvement is ``in there'', but the interesting question is: ``how do we build the guide that finds it?''

We therefore propose three directions, each expanding the role of the guide.
(i) the guide as a design object, with three levels of intervention when the diagnostics fail: recalibrating a guide whose ranking survives, refitting a guide whose discrimination has failed, as on Maze-Hard (Appendix~\ref{app:guide_quality}), and, where neither suffices, bespoke guiding, a guide built for the task to handle structural constraints, custom training objectives or multi-guide compositions rather than inherited from the $Q$-head;
(ii) train-time ramifications of the theory (Appendix~\ref{sec:app_checkpoint}) and inference-time efficiency, from across-particle parallelism (Remark~\ref{rem:deterministic_limit}) to scaling beyond Figure~\ref{fig:ablation_sensitivity}; and (iii) broader evaluation beyond pencil puzzles, from graph-based \citep{palm2018recurrent,sinha2019clutrr} and rule-governed language tasks \citep{clark2020transformers,tafjord2021proofwriter} to ARC-AGI \citep{arcprize2025arc2}.

More broadly, we propose generalisations of the framework beyond recursive reasoning. In AI safety, steering has emerged as a promising approach for online adaptation of foundation-model behaviour \citep{turner2023steering,rimsky2024steering}; a guided variant brings an explicit mechanism for control, in which a bespoke guide identifies sampled trajectories aligned with pre-trained constraints rather than rotating activation vectors directly.

A natural next question is the transfer-learning capacity of small recursive models. If a guide $q_A$ is used up to step $n$ to produce $(y_n,z_n)$, can $z_n$ seed another task, or can a second guide $q_B$ redirect the same state towards a related solution? This question is especially natural for graph-based reasoning \citep{palm2018recurrent,sinha2019clutrr} and rule-governed language tasks \citep{clark2020transformers,tafjord2021proofwriter}, where success is determined by consistent relational or logical constraints. If such transfer is proven possible, tiny recursive models would become more than compact solvers: they would be reusable reasoning devices, generalist in capability and specialised online with a guide.

\section*{Acknowledgements and disclosure of funding}
This work was funded by digiLab. The authors declare no competing interests.

\bibliography{refs}
\bibliographystyle{plainnat}

\appendix

\section{Evaluation protocol}
\label{sec:app_eval_protocol}

\paragraph{Models and data.}
We consider two frozen TRM models trained with the Nano-TRM framework~\citep{koch2025nano_trm} on Sudoku-Extreme and Maze-Hard~\citep{wang2025hrm}, using Nano-TRM's default training hyperparameters, and a third that we train on PPBench (Appendix~\ref{app:ppbench}). Following the data pipeline released with \citet{wang2025hrm}, we apply $1{,}000$ shuffling augmentations to Sudoku-Extreme training data and no augmentation to Maze-Hard. The model, dataset, and outer horizon $N=48$ are held fixed throughout, matching \citet{jolicoeur2025trm}, whose $16$ supervision steps of $3$ deep recursions give the same $48$ outer steps, so only the inference procedure (\S\ref{sec:method}) varies.
Figure~\ref{fig:maze_case_study} (\S\ref{app:maze_case}) shows the contrasting training trajectories for the two $Q$-heads.

\paragraph{Splits and statistics.}
For Sudoku-Extreme, we hold out a $500$-puzzle validation subset for hyperparameter selection, disjoint from both the $10{,}020$-puzzle Test split and the $1{,}427$-puzzle deterministic-failure (DF) split (DF puzzles are those on which the deterministic TRM baseline fails). The seed-fold runs of Table~\ref{tab:main_results} evaluate $9{,}984$ Test and $1{,}408$ DF puzzles; the full-set trace analyses in the appendices evaluate every puzzle of both splits.
For Maze-Hard, the $1{,}000$-puzzle Test split (of which the seed-fold runs evaluate $992$ puzzles and the trace analyses all $1{,}000$) is too small to subdivide; no validation subset is held out, and hyperparameters are transferred from the Sudoku-Extreme sweep (discussion below).
Headline experiments are run over $5$ data folds (disjoint partition of Test, each fold $20\%$ of Test; see Table~\ref{tab:hyperparameters}) $\times\,5$ random seeds (initialisation of stochastic variables), giving $25$ independent runs; reported $\pm$ values are seed-fold standard deviations.
Hyperparameter ablations on Sudoku-Extreme (Figure~\ref{fig:ablation_sensitivity}) use a $2{,}000$-puzzle subset of Test for compute reasons.

\paragraph{Puzzles evaluated and call budgets.}
Table~\ref{tab:puzzle_counts} lists, for each benchmark split, how many puzzles each procedure evaluated and its backbone calls per puzzle. The core runs evaluated slightly fewer puzzles than the splits contain; the stored-trace analyses of Appendices~\ref{app:selection_rules} to~\ref{app:depth_cost} evaluate every puzzle. The PPBench core runs are $5$ seeds of one fold.

\begin{table}[h]
  \centering
  \small
  \caption{\textbf{Puzzles evaluated and per-puzzle call budgets.} Backbone calls are $S\times N\times(M+1)$ for the particle filter and $N\times(M+1)$ for the deterministic TRM, both at $N=48$.}
  \label{tab:puzzle_counts}
  \setlength{\tabcolsep}{4pt}
  \begin{tabular}{@{}llccccc@{}}
    \toprule
    Benchmark & Split & In split & Core runs & Stored-trace analyses & Filter calls & Deterministic calls \\
    \midrule
    Sudoku-Extreme & Test & $10{,}020$ & $9{,}984$ & $10{,}020$ & $5{,}376$ & $336$ \\
    Sudoku-Extreme & DF & $1{,}427$ & $1{,}408$ & $1{,}427$ & $5{,}376$ & $336$ \\
    Maze-Hard & Test & $1{,}000$ & $992$ & $1{,}000$ & $3{,}840$ & $240$ \\
    Maze-Hard & DF & $134$ & $128$ & $134$ & $3{,}840$ & $240$ \\
    PPBench & Validation & $253$ & $253$ & $253$ & $3{,}840$ & $240$ \\
    PPBench & Golden & $42$ & $42$ & $42$ & $3{,}840$ & $240$ \\
    \bottomrule
  \end{tabular}
\end{table}

\paragraph{Compute resources.}
All training and inference runs use a single NVIDIA H100 GPU. The full set of reported experiments (headline $5\times 5$ seed-fold sweeps on Sudoku-Extreme and Maze-Hard, hyperparameter ablations, the checkpoint scan, and figure-data generation) consumes approximately $20$--$30$ GPU-hours in total. Preliminary or discarded runs that did not make it into the final paper are not included in this estimate but are of comparable order. The PPBench backbone training, its five-seed inference runs and the stored-trace analyses of Appendices~\ref{app:selection_rules} to~\ref{app:depth_cost} were run afterwards on the same hardware.

\paragraph{Implementation and cost.}
Weights are carried in log space and normalised by log-sum-exp; the ESS and the systematic resampling are computed from the normalised weights. Table~\ref{tab:cost} reports batched throughput and peak memory on one NVIDIA H100 at $N=48$: $64$ puzzles in four batches of $16$ per setting, timed end to end, so the figures are amortised seconds per puzzle rather than single-query latency. Memory grows linearly in $S$ (the state is $2LD$ floats per particle); wall-clock is nearly flat to $S=8$ and roughly four times the single-particle cost at $S=16$.

\begin{table}[h]
  \centering
  \small
  \caption{\textbf{Measured inference cost against particle count} ($N=48$, one NVIDIA H100, $64$ puzzles per setting).}
  \label{tab:cost}
  \begin{tabular}{@{}lccccc@{}}
    \toprule
    $S$ & $1$ & $2$ & $4$ & $8$ & $16$ \\
    \midrule
    Seconds per puzzle & $0.014$ & $0.013$ & $0.018$ & $0.031$ & $0.058$ \\
    Peak GPU memory (MB) & $84$ & $111$ & $185$ & $307$ & $554$ \\
    \bottomrule
  \end{tabular}
\end{table}

\begin{table*}[h]
  \centering
  \small
  \caption{\textbf{Hyperparameters used for main results.} Common across all three benchmarks: $S=16$, $N=48$; $\tau_{\ESS}=0.3$ on Sudoku-Extreme and Maze-Hard and $0.5$ on PPBench, fixed a priori. The noise scale $\sigma$, inverse temperature $\beta$ and trained inner depth $M$ vary by benchmark; selection methodology in `Hyperparameter selection', \S\ref{sec:app_eval_protocol}.}
  \vspace{1em}
  \label{tab:hyperparameters}
  \begin{tabular}{@{}lcccccc@{}}
    \toprule
    Model & $\sigma$ & $\beta$ & $M$ & Splits & Design & Reported as \\
    \midrule
    Sudoku-Extreme & $0.3$ & $0.25$ & $6$ & Test ($9{,}984$), DF ($1{,}408$) & $5\!\times\!5$ seed-fold & $\pm$ seed--fold s.d.\\
    Maze-Hard      & $0.5$ & $1$ & $4$ & Test ($992$)                & $5\!\times\!5$ seed-fold    & $\pm$ seed--fold s.d.\\
    PPBench        & $0.3$ & $0.25$ & $4$ & Val ($253$), Golden ($42$) & $5$ seeds, one fold & $\pm$ seed s.d.\\
    \bottomrule
  \end{tabular}
\end{table*}

\paragraph{Hyperparameter selection.}
Three hyperparameters are common to Sudoku-Extreme and Maze-Hard: $S=16$ particles, $N=48$ outer steps, $\tau_{\ESS}=0.3$; PPBench reuses $S$, $N$, $\sigma$ and $\beta$ from the Sudoku-Extreme sweep with $\tau_{\ESS}=0.5$ fixed a priori (Appendix~\ref{app:ppbench}). Only the noise scale $\sigma$ and inverse temperature $\beta$ vary by benchmark (Table~\ref{tab:hyperparameters}); the inner depth $M$ is the checkpoint's trained value ($6$ and $4$) and is not a free parameter. For Sudoku-Extreme, we sweep $\mathfrak{H} := (\sigma, \beta, S, \tau_{\mathrm{ESS}})$ on the validation subset and select $(\sigma, \beta) = (0.3, 0.25)$; reported Test and DF numbers are then a single-shot evaluation of this configuration.
For Maze-Hard, no validation set is available, and we deliberately do not tune on Maze-Hard data: $S$, $N$, $\tau_{\ESS}$ are transferred from the Sudoku-Extreme sweep, $\sigma=0.5$ follows the architectural prior (Maze-Hard's normalisation scale is larger than Sudoku-Extreme's), and $\beta=1$ is a canonical default for non-trivial tempering, rather than the more conservative $\beta=0.25$ used for Sudoku-Extreme. The role of Maze-Hard here is to test whether the diagnostics of \S\ref{sec:theory} correctly predict failure on a configuration chosen without Maze-Hard data, which they do without observing ground truth data (Appendix~\ref{app:maze_case}).

\paragraph{Ablation scope.}
The hyperparameter sweeps in \S\ref{sec:app_ablation} are performed on the held-out Test sets and serve as post-tuning robustness checks for the validated (Sudoku-Extreme) and diagnostic-selected (Maze-Hard) configurations. Consistent with the diagnostic prediction, the Maze-Hard $Q$-head induces a constant response to $\beta$ (Figure~\ref{fig:ablation_sensitivity}), confirming that no $\beta$ sweep recovers the headroom.

\section{Maze-Hard: further analysis}
\label{app:maze_case}
\begin{figure*}[h]
  \centering
    \includegraphics[width=\textwidth]{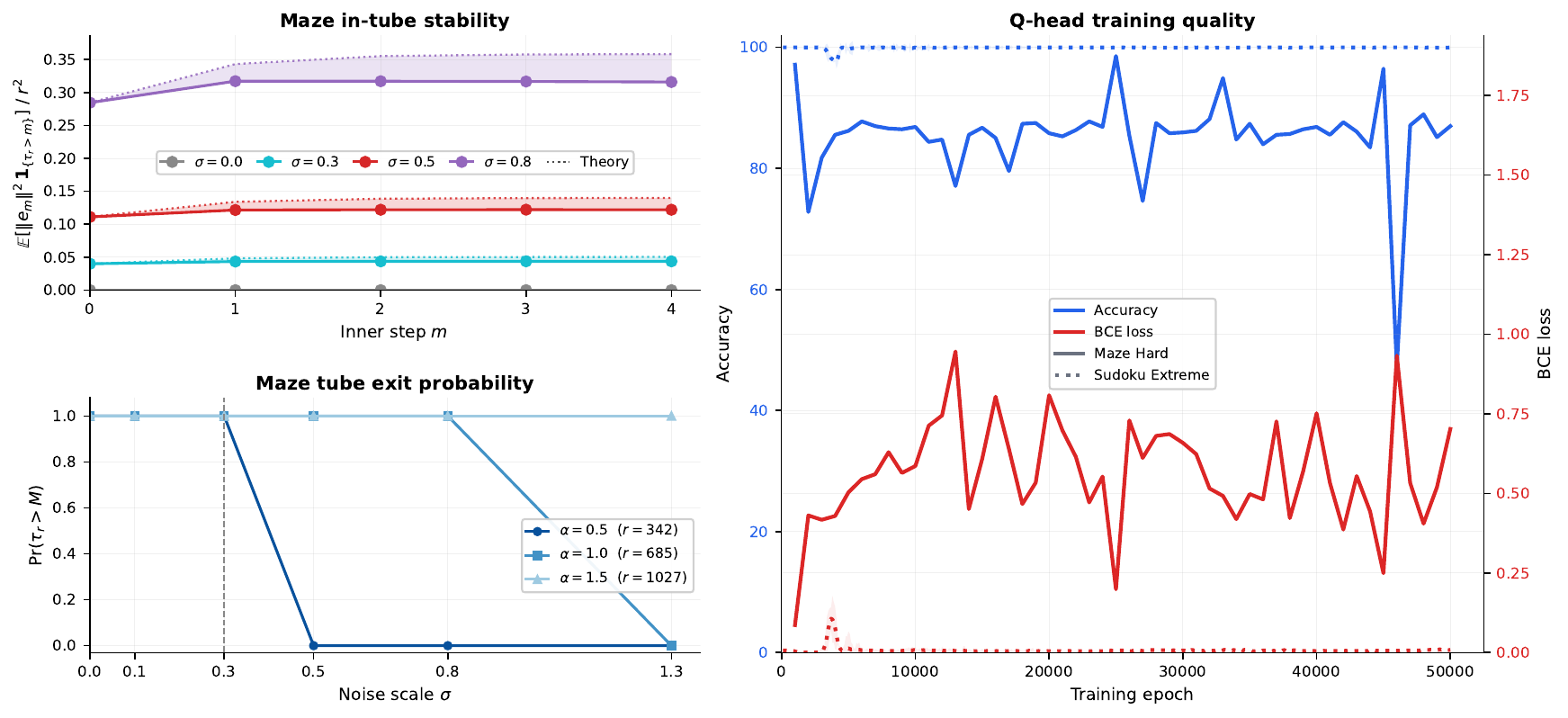}
    \caption{\textbf{Maze-Hard case study.} \textit{Top, left}: In-tube deviations $\mathbb{E}[\|e_m\|^2\mathbf{1}_{\{\tau_r>m\}}]/r^2$ per inner step $m$ at $\alpha=1.5$, with upper bounds (dotted) from the empirical local-Lipschitz estimate $\hat\rho_r$ defined in Figure~\ref{fig:inner_deviation}. \textit{Bottom, left}: Tube-survival probability $\Pr(\tau_r>M)$ vs.\ $\sigma$ for $r=\alpha\sqrt{LD}$; stable for $\alpha\ge 0.5$, $\sigma\le 0.3$, matching \eqref{eq:tube_pre_exit_bound}. \textit{Right}: $Q$-head halting accuracy and BCE loss on the training framework's evaluation split, as logged during training. Sudoku-Extreme saturates near zero BCE and $100\%$ accuracy, while Maze-Hard fails to converge.}
  \label{fig:maze_case_study}
\end{figure*}

Figure~\ref{fig:maze_case_study} places the Maze-Hard tube-stability evidence, the analogue of Figure~\ref{fig:inner_deviation} for Sudoku-Extreme, alongside the $Q$-head training diagnostics for both models.
Reading the \textit{left} panel shows that, much like Figure \ref{fig:inner_deviation}, the stochastic trajectories of the Maze-Hard model are stable in the sense of Proposition~\ref{prop:tube_stability} for $\alpha\ge 0.5$ and $\sigma\le 0.3$; the range of stable $\sigma$ values conforms to the in-tube stability bound \eqref{eq:tube_pre_exit_bound}. However, the \textit{right} panel shows what we believe is a justification for the poor performance of the in-built guide, observed in Figure~\ref{fig:guide_alignment}. We display the training curves for the $Q$-head on both models, showing extremely high accuracy in the Sudoku-Extreme case, and poor, highly unstable convergence on Maze-Hard. We conjecture this is in part due to the sensitivity of the Maze-Hard structure, in that minor deviations from the solutions lead to poor results, given the objective is to find the `shortest' path to the goal, not just an admissible one; this bimodal objective is not well supported by the BCE objective.
A constructive use of the diagnostics for hyperparameter selection (rather than verification) is a natural follow-up; in this paper they serve only as a label-free verifier.

\section{Hyperparameter ablation analysis}
\label{sec:app_ablation}

\begin{figure*}[h]
  \centering
  \includegraphics[width=\textwidth]{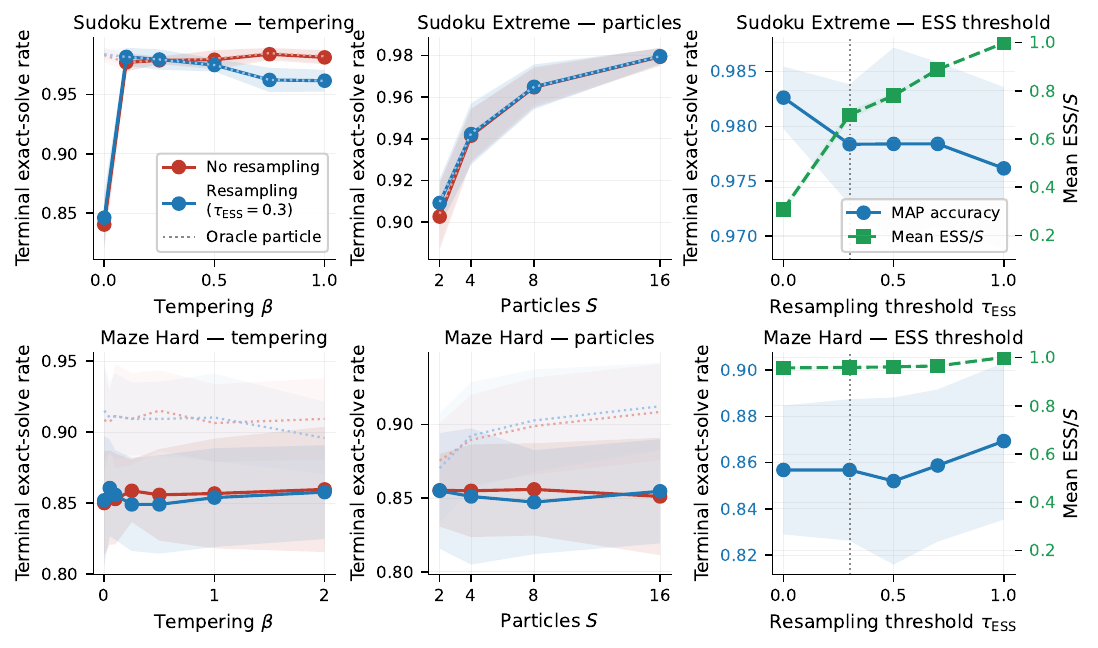}
  \caption{\textbf{Hyperparameter sensitivity sweeps.} We report sweeps for Sudoku-Extreme (\textit{top}) and Maze-Hard (\textit{bottom}). \textit{Left}: we vary $\beta$ across scales, with and without adaptive resampling. \textit{Middle}: we vary the number of particles in the cloud, $S\in\{2, 4, 8, 16\}$. \textit{Right}: we vary the resampling threshold $\tau_{\mathrm{ESS}}$. All other hyperparameters are fixed to the values in Table~\ref{tab:hyperparameters}.}
  \label{fig:ablation_sensitivity}
\end{figure*}

We report hyperparameter sweeps that complement the headline results of \S\ref{sec:experiments}; Figure~\ref{fig:ablation_sensitivity} summarises the observations across both benchmarks.

\paragraph{Sudoku-Extreme (top).}
The $\beta$ sweep shows that even small amounts of $Q$-guidance recover the oracle particle, consistent with the high training-time accuracy of the $Q$-head (Figure~\ref{fig:maze_case_study}, \textit{right}). Once the oracle is recovered, larger $\beta$ has no further upside; resampling at high $\beta$ only distorts the cloud weighting. Terminal accuracy increases with the particle count $S$ (\textit{middle}). This is encouraging for scaling, and motivates us to consider adaptive particle-count schemes as a follow-up.

\paragraph{Maze-Hard (bottom).}
The sweeps tell quite a clear story: the guide is ineffective at identifying successful trajectories in any regime. The oracle gap is consistently visible, i.e. the cloud contains successful trajectories for every setting, but none of $\beta$, $S$, or $\tau_{\mathrm{ESS}}$ has any effect on MAP performance. It is noteworthy that the normalised effective sample size holds at $\mathrm{ESS}_n/S \approx 1$ throughout the sweeps (meaning weights stay uniform; no reweighting takes place) so the collapse is a property of the trained $Q$-head, not of the inference procedure. This makes sense given the near-zero variance in $q_\theta$ values (Figure~\ref{fig:guide_alignment}, \textit{bottom}).

\paragraph{Inner depth and noise family.}
Two further sweeps, each on $5$ folds of one seed, test the two choices that the main results hold fixed. Table~\ref{tab:inner_depth} varies the inner depth $M$ at inference on a $1{,}008$-puzzle Sudoku-Extreme Test subset with $S=16$ and $N=48$: cutting the trained depth to $M=2$ costs $17$ points and the oracle falls with it, so the loss is the backbone's rather than the guide's, while $M\ge4$ is flat within a fold s.d. We therefore keep the trained $M$ ($6$ on Sudoku-Extreme, $4$ on Maze-Hard and PPBench) and control exploration through $\sigma$. Table~\ref{tab:noise_families} replaces the Gaussian proposal noise by Laplace or Student-$t$ noise, each scaled to unit variance per coordinate so that $\nu^2$ in Proposition~\ref{prop:tube_stability} is unchanged (Remark~\ref{rem:noise_assumptions}). Accuracy and oracle move by less than a fold s.d.\ on both benchmarks; the guide spread is unchanged on Sudoku-Extreme and rises from $0.23$ to $0.33$ under Student-$t$ noise on Maze-Hard, still five times below the Sudoku-Extreme value, so the Maze-Hard guide stays flat under every family.

\begin{table}[h]
  \centering
  \small
  \caption{\textbf{Inner depth at inference} on a $1{,}008$-puzzle Sudoku-Extreme Test subset ($S=16$, $N=48$; exact-solve \%, mean $\pm$ s.d.\ over $5$ folds of one seed). The trained depth is $M=6$; its full-split values are in Table~\ref{tab:main_results}.}
  \label{tab:inner_depth}
  \begin{tabular}{@{}lcccc@{}}
    \toprule
    & $M=2$ & $M=4$ & $M=8$ & $M=12$ \\
    \midrule
    Sequential guiding (ours) & $81.1\!\pm\!3.5$ & $97.9\!\pm\!1.3$ & $97.7\!\pm\!0.8$ & $98.4\!\pm\!0.7$ \\
    Oracle (guided cloud) & $84.0\!\pm\!1.9$ & $98.0\!\pm\!1.1$ & $97.7\!\pm\!0.8$ & $98.4\!\pm\!0.7$ \\
    \bottomrule
  \end{tabular}
\end{table}

\begin{table}[h]
  \centering
  \small
  \caption{\textbf{Proposal noise family} (exact-solve \%, mean $\pm$ s.d.\ over $5$ folds of one seed; Sudoku-Extreme on a $1{,}008$-puzzle Test subset, Maze-Hard on the full $1{,}000$-puzzle Test split). Laplace noise has scale $1/\sqrt2$ and the Student-$t$ noise has three degrees of freedom scaled by $1/\sqrt3$, so every family has unit variance per coordinate. Guide spread is the s.d.\ of the terminal raw logit over particles. The Gaussian configuration was not rerun on the Sudoku-Extreme subset; its full-split values are in Table~\ref{tab:main_results}.}
  \label{tab:noise_families}
  \begin{tabular}{@{}llccc@{}}
    \toprule
    & Family & Sequential guiding (ours) & Oracle (guided cloud) & Guide spread \\
    \midrule
    Sudoku-Extreme & Laplace & $98.0\!\pm\!1.4$ & $98.0\!\pm\!1.4$ & $1.737\!\pm\!0.052$ \\
     & Student-$t$ & $98.2\!\pm\!0.7$ & $98.3\!\pm\!0.9$ & $1.757\!\pm\!0.164$ \\
    \midrule
    Maze-Hard & Gaussian & $85.6\!\pm\!4.4$ & $90.9\!\pm\!3.3$ & $0.234\!\pm\!0.012$ \\
     & Laplace & $85.5\!\pm\!4.1$ & $91.3\!\pm\!3.1$ & $0.231\!\pm\!0.016$ \\
     & Student-$t$ & $84.8\!\pm\!4.6$ & $91.7\!\pm\!2.8$ & $0.327\!\pm\!0.016$ \\
    \bottomrule
  \end{tabular}
\end{table}

\section{Trajectory contraction in token entropy}
\label{sec:contraction}

To understand the capacity for calibrated uncertainty quantification a little further, we ask the question: `how does accumulated entropy measure up against final-prediction entropy?' The answer, we propose, sheds light on the evolution of uncertainty along the reasoning trajectories.

Specifically, we introduce a blunt measure of the relationship between the two statistics introduced in \S\ref{sec:token_entropy}: the terminal $H_\ell$ \eqref{eq:token_entropy_terminal} and the path-integrated $\bar H_\ell$ \eqref{eq:token_entropy_aggregated}. Define
\begin{equation}
C_\ell := \bar H_\ell - H_\ell,
\label{eq:contraction}
\end{equation}
the per-token \textit{contraction} between the trajectory mean and the terminal snapshot. Positive $C_\ell$ therefore implies that the token marginal sharpened towards the final prediction.

\begin{figure*}[h]
  \centering
  \includegraphics[width=\linewidth]{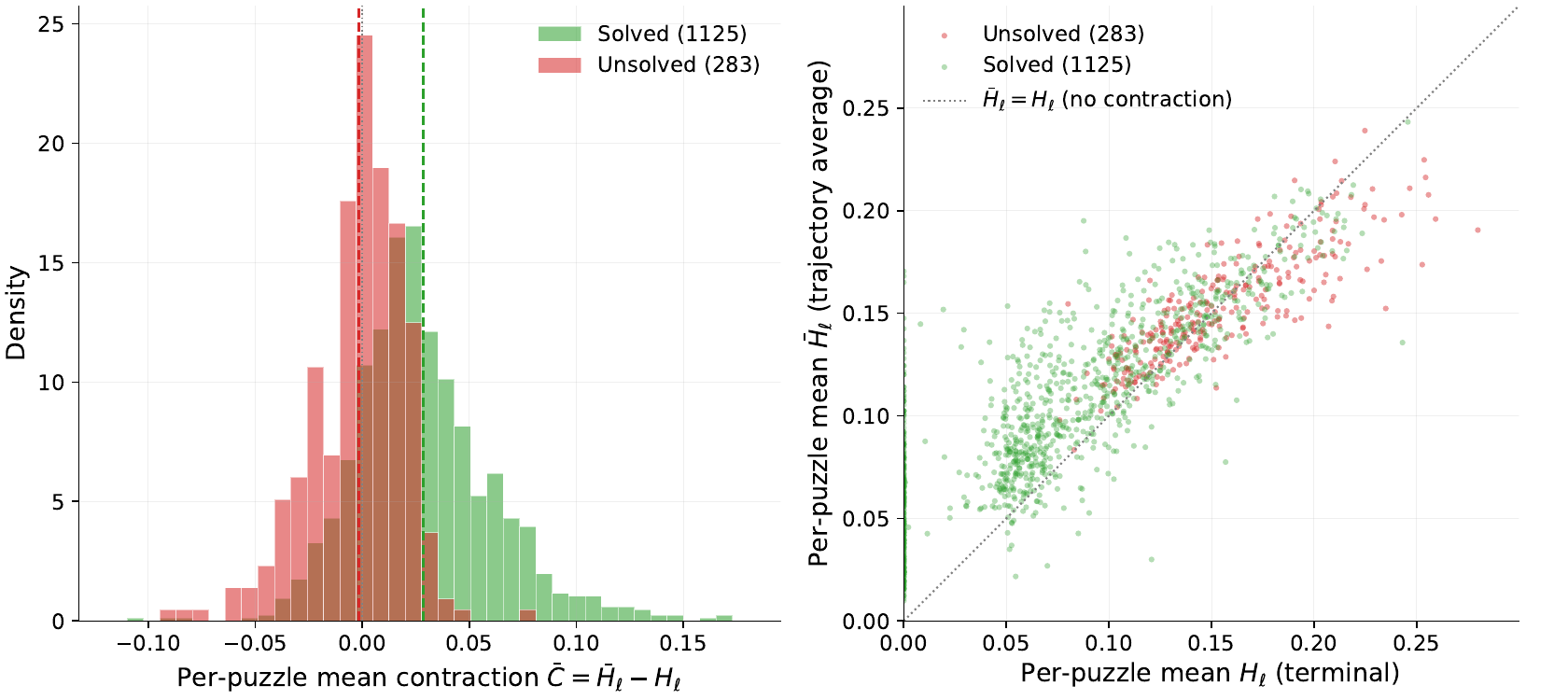}
  \caption{\textbf{Trajectory contraction on the deterministic-failure split.} \textit{Left}: histogram of per-puzzle mean contraction $\bar C := \text{mean}_\ell C_\ell$ over unmasked tokens, indexed by guided MAP solve outcome. Solved puzzles have mean $0.028$ whilst unsolved puzzles have mean $-0.002$. \textit{Right}: per-puzzle scatter of trajectory-averaged mean entropy $\bar H_\ell$ against terminal mean entropy $H_\ell$, with the identity, $C_{\ell}=0$, shown.}
  \label{fig:token_contraction}
\end{figure*}

Figure~\ref{fig:token_contraction} reports the per-puzzle average of $C_\ell$. The histograms (\textit{left}), separated by solve outcome, reinforce the utility of the diagnostic. Correctly solved puzzles, i.e. every token is correct, admit a clear positive contraction, as expected.

The more remarkable feature is the negative skew of the unsolved-puzzle histogram. With mean contraction near zero, the long left tail represents deepening entropy with the token marginal becoming less concentrated towards the terminal step. This implies that the trajectories continue to explore rather than settle on a stable incorrect solution.

The scatter plot (\textit{right}) gives a bird's-eye view of the same data, exposing the magnitudes of $\bar H_\ell$ and $H_\ell$ rather than only their difference. As expected from Figure~\ref{fig:token_entropy} (\textit{right}), the highest entropy clusters among unsolved puzzles. More striking is the threshold at roughly $(0.1, 0.1)$: below it, every prediction in the cloud is correct. Could these entropy statistics serve as calibrated guides in their own right?

\section{Sudoku-Extreme: train-time efficiency}
\label{sec:app_checkpoint}

\begin{figure*}[h]
  \centering
    \includegraphics[width=0.95\linewidth]{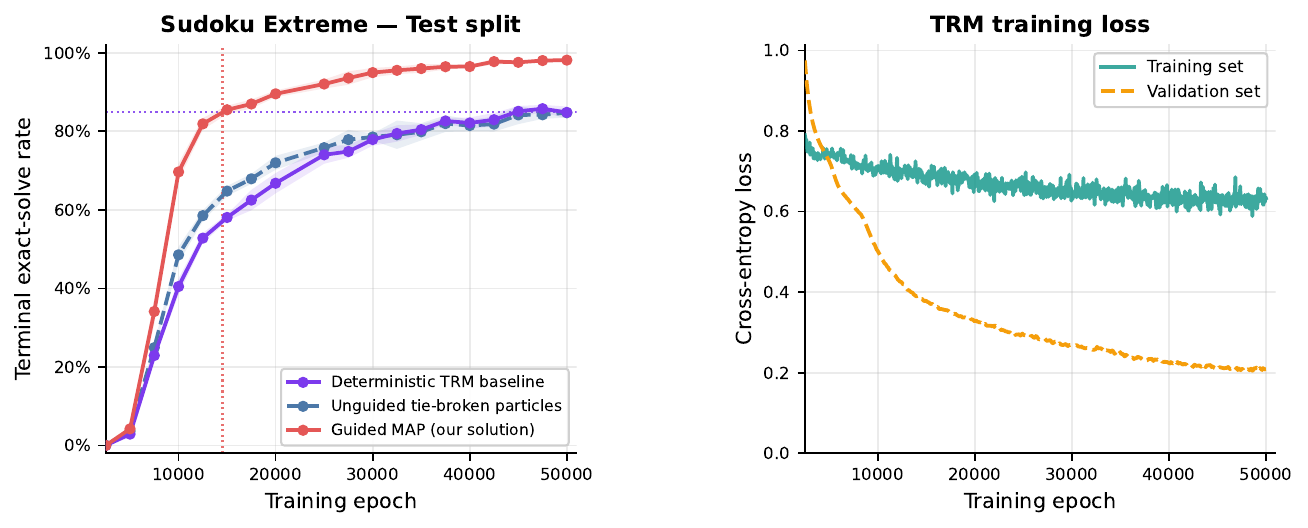}
    \caption{\textbf{Per-checkpoint scan over the TRM training run.} \textit{Left}: terminal exact-solve rate of guided MAP (red), unguided MAP (blue), and deterministic TRM (purple) across $19$ training checkpoints (epochs $2{,}499$--$49{,}999$) on the Sudoku-Extreme test split; bands are five-fold standard deviations. Dotted horizontal: deterministic terminal accuracy at epoch $49{,}999$. Dotted vertical: epoch at which guided MAP first crosses this threshold ($\approx 3.1\times$ earlier than the deterministic readout). \textit{Right}: TRM training-loss trace (train and validation token cross-entropy) over the same epochs.}
  \label{fig:checkpoint_scan}
\end{figure*}

In this section we ask whether guided reasoning acts as a train-time accelerator: how much earlier in training can it reach the accuracy achieved by the deterministic readout at convergence? Figure~\ref{fig:checkpoint_scan} applies Algorithm~\ref{alg:guided_inference} to $19$ stored checkpoints from the Sudoku-Extreme training run (epochs $2{,}499$--$49{,}999$), under the same five-fold evaluation as the main results; the training loss is shown on the \textit{right}.

The \textit{left} panel reveals two observations. First, guided MAP reaches the deterministic terminal accuracy approximately $3.1\times$ earlier in training than the deterministic readout itself. We take this as a meaningful training-economics signal.
Second, the method is stable across the full checkpoint range, with no fragility around any particular epoch; this also serves as a secondary ablation on the checkpoint stability of the deterministic TRM core itself.

\begin{remark}
The $Q$-head is far cheaper to train than the TRM backbone: it is a classifier whose loss converges quickly (Figure~\ref{fig:maze_case_study}, \textit{right}), whereas the backbone is the expensive object (Figure~\ref{fig:checkpoint_scan}, \textit{right}). This asymmetry supports the bespoke-guide direction in \S\ref{sec:discussion}: a separately trained guide can ride on top of an existing TRM without re-touching the backbone.
\end{remark}

\section{Analysis of the class-conditional gap}
\label{sec:app_gap}

In this section we dig deeper into the guide alignment diagnostic of \S\ref{sec:bce_alignment}.
We specialise the Gibbs covariance identity \eqref{eq:gibbs_covariance} to the binary-success case, yielding a log-odds derivative that directly links the class-conditional gap $\Delta_n$ to the mass shift on the success set.

Fix an outer depth $n$ and a measurable success set $A_n \subseteq \mathcal H$; in our experiments, $A_n$ is the set of states that solve the puzzle by the terminal horizon $N$. Write $\varphi := \mathbf 1_{A_n}$ and $p_n(\beta) := \mathbb P_{\pi_{n+1,\beta}}(H \in A_n)$. Applying \eqref{eq:gibbs_covariance} with $\varphi = \mathbf 1_{A_n}$ and expanding the covariance gives
\begin{equation}
\frac{\mathrm d}{\mathrm d\beta}\, p_n(\beta)
= \operatorname{Cov}_{\pi_{n+1,\beta}}\!\bigl(\mathbf 1_{A_n},\,q_\theta\bigr)
= p_n(\beta)\bigl(1-p_n(\beta)\bigr)\,\Delta_n(\beta),
\label{eq:binary_derivative}
\end{equation}
where (cf.\ \eqref{eq:delta})
\begin{equation}
\Delta_n(\beta)
=
\mathbb E_{\pi_{n+1,\beta}}\!\bigl[q_\theta(H)\mid H\in A_n\bigr]
-
\mathbb E_{\pi_{n+1,\beta}}\!\bigl[q_\theta(H)\mid H\in A_n^c\bigr]
\end{equation}
is the class-conditional mean gap in guide score between success and failure states. For $0<p_n(\beta)<1$, applying the logit chain rule
$\tfrac{\mathrm d}{\mathrm d\beta}\log\tfrac{p}{1-p}=\tfrac{1}{p(1-p)}\tfrac{\mathrm d p}{\mathrm d\beta}$
to \eqref{eq:binary_derivative} cancels the variance factor and yields
\begin{equation}
\frac{\mathrm d}{\mathrm d\beta}\log\frac{p_n(\beta)}{1-p_n(\beta)}
=
\Delta_n(\beta).
\label{eq:logodds_derivative}
\end{equation}

Equation~\eqref{eq:logodds_derivative} is the labelled form of the one-number test in \S\ref{sec:bce_alignment}: tempering strictly improves the log-odds of success whenever $\Delta_n(\beta)>0$, and cannot help when the two classes are indistinguishable under $q_\theta$. The general alignment notion of \S\ref{sec:bce_alignment} reduces here to $\Delta_n(\beta) \ge 0$ along the tilting path. The binary-success case is the natural setting for future custom guides; for example, classifiers identifying answers that lie outside a pre-trained constraint set (\S\ref{sec:discussion}), or those that forecast future success (via a delayed training reward).

\section{Proofs supporting \S\ref{sec:theory}}
\label{app:proofs}

Here we record the proofs of the theoretical results stated in \S\ref{sec:theory}: the BCE-alignment lemma (Lemma~\ref{lem:bce_aligned}, \S\ref{sec:bce_alignment}) for any strictly increasing transformation, the tube-stability bound (Proposition~\ref{prop:tube_stability}, \S\ref{sec:local_stability}), and the resulting tube-exit probability (Corollary~\ref{cor:tube_exit}).

\subsection{Proof of Lemma~\ref{lem:bce_aligned}}
\label{sec:app_proof_of_lemma}

The tilt $\exp(t\,\psi(\eta))$ depends on $h$ only through $\eta$, so conditioning on $\eta$ removes it: the conditional law of $Y$ given $\eta$ is the same under every $\pi_{n+1,t}$, $t\in[0,\beta]$, and the hypothesis $\mathbb E[Y\mid\eta]=\eta$ holds under each. By the tower property it gives $\mathbb E[Y\,\psi(\eta)]=\mathbb E[\eta\,\psi(\eta)]$ and $\mathbb E[Y]=\mathbb E[\eta]$ under $\pi_{n+1,t}$, hence
\begin{equation}
  \operatorname{Cov}_{\pi_{n+1,t}}\!\bigl(Y,\,\psi(\eta)\bigr)
  \;=\;
  \operatorname{Cov}_{\pi_{n+1,t}}\!\bigl(\eta,\,\psi(\eta)\bigr).
  \label{eq:tower_reduction}
\end{equation}
Strict monotonicity of $\psi$ gives $(\eta-\eta')(\psi(\eta)-\psi(\eta'))\ge 0$ pointwise for an i.i.d.\ pair $(\eta,\eta')$, and the standard identity $2\operatorname{Cov}(U,V)=\mathbb E[(U-U')(V-V')]$ for i.i.d.\ copies gives
\begin{equation}
  2\operatorname{Cov}\bigl(\eta,\,\psi(\eta)\bigr)
  =
  \mathbb E\bigl[(\eta-\eta')(\psi(\eta)-\psi(\eta'))\bigr]
  \;\ge\;0,
  \label{eq:cov_via_iid_pair}
\end{equation}
with equality iff $\eta$ is constant on the support of $\pi_{n+1,t}$. The case $\psi=\log$ gives the BCE-population statement. \qed

\subsection{Proof of Proposition~\ref{prop:tube_stability}}

Fix $h_n=(y_n,z_n)$ and suppress $n$, $x$, $h_n$ from the notation, setting
\begin{equation}
F_{\theta}(z):=f_\theta(x+y_n+z).
\end{equation}
Then the deterministic and stochastic inner trajectories satisfy $z_1:=z_{n,1} = f_\theta(x+y_n+z_n)$, $z_{m+1}=F_{\theta}(z_m)$, as well as $\zeta_1=z_1+\sigma\xi_0$, $\zeta_{m+1}=F_{\theta}(\zeta_m)+\sigma\xi_m$ for $m=1,\dots,M-1$.

Write $e_m:=\zeta_m-z_m$ as in \eqref{eq:em_taur}, and for $m\ge 0$ let
\begin{equation}
E_m:=\{\|e_1\|\le r,\dots,\|e_m\|\le r\},
\label{eq:em_exit}
\end{equation}
with the convention $E_0 := \Omega$ (the empty intersection); equivalently, $E_m = \{\tau_r > m\}$ for $m \ge 0$.

\textit{Base case.} Since $e_1=\sigma\xi_0$ and $\mathbf 1_{E_1}\le 1$,
\begin{equation}
\mathbb{E}\bigl[\|e_1\|^2\mathbf 1_{E_1}\bigr] \le \sigma^2\mathbb{E}\|\xi_0\|^2 \le \sigma^2\nu^2.
\label{eq:proof_base}
\end{equation}

\textit{One-step recursion.} Fix $m\in\{1,\dots,M-1\}$, and write
\begin{equation}
D_m:=F_{\theta}(\zeta_m)-F_{\theta}(z_m),
\label{eq:am}
\end{equation}
so that $e_{m+1}=D_m+\sigma\xi_m$. Condition on $\mathcal F_m:=\sigma(\xi_0,\dots,\xi_{m-1})$: both $D_m$ and $\mathbf 1_{E_m}$ are $\mathcal F_m$-measurable, while $\xi_m\perp\mathcal F_m$ has mean zero. Expanding the square and taking expectations, the cross-term vanishes and the noise term satisfies $\mathbb{E}[\|\xi_m\|^2\mathbf 1_{E_m}]\le\mathbb{E}\|\xi_m\|^2\le\nu^2$, so
\begin{equation}
\mathbb{E}\left[\|e_{m+1}\|^2\mathbf 1_{E_m}\right]
\le
\mathbb{E}\left[\|D_m\|^2\mathbf 1_{E_m}\right]+\sigma^2\nu^2.
\label{eq:proof_cross_split}
\end{equation}

On $E_m$, $\|e_m\|\le r$, so $\zeta_m\in\{z:\|z-z_m\|\le r\}$; both $z_m$ and $\zeta_m$ therefore lie in $\bigcup_{j=1}^{M}\{z:\|z-z_j\|\le r\}$, and the local Lipschitz hypothesis gives $\|D_m\|\le \rho_r\|e_m\|$ on $E_m$. Substituting into~\eqref{eq:proof_cross_split} yields the one-step bound
\begin{equation}
\mathbb{E}\left[\|e_{m+1}\|^2\mathbf 1_{E_m}\right]
\le
\rho_r^2\,\mathbb{E}\left[\|e_m\|^2\mathbf 1_{E_m}\right]+\sigma^2\nu^2.
\label{eq:proof_one_step}
\end{equation}
Since $E_{m+1}\subseteq E_m$, in particular
\begin{equation}
\mathbb{E}\left[\|e_{m+1}\|^2\mathbf 1_{E_{m+1}}\right]
\le
\rho_r^2\,\mathbb{E}\left[\|e_m\|^2\mathbf 1_{E_m}\right]+\sigma^2\nu^2.
\label{eq:proof_bound3}
\end{equation}

\textit{Iterating the recursion.} Define $a_m:=\mathbb{E}\left[\|e_m\|^2\mathbf 1_{E_m}\right]$ for $m=1,\dots,M$. By \eqref{eq:proof_base} and \eqref{eq:proof_bound3},
\begin{equation}
a_1\le \sigma^2\nu^2,
\qquad
a_{m+1}\le \rho_r^2 a_m+\sigma^2\nu^2.
\end{equation}
Iterating gives
\begin{equation}
a_m
\le
\sigma^2\nu^2\sum_{j=0}^{m-1} \rho_r^{2j}
\le
\frac{\sigma^2\nu^2}{1-\rho_r^2}
\end{equation}
for $m=1,\dots,M$, which recovers \eqref{eq:tube_pre_exit_bound}.
\qed

\subsection{Proof of Corollary~\ref{cor:tube_exit}}

Using the notation and conventions from the previous proof ($E_m=\{\tau_r>m\}$ for $m\ge 0$, with $E_0=\Omega$), define
\begin{equation}
b_m:=\mathbb{E}\left[\|e_m\|^2\mathbf 1_{E_{m-1}}\right]
\qquad(m=1,\dots,M).
\end{equation}
We first show that
\begin{equation}
b_m
\le
\sigma^2\nu^2\sum_{j=0}^{m-1} \rho_r^{2j}
\le
\frac{\sigma^2\nu^2}{1-\rho_r^2}.
\label{eq:bm_bound}
\end{equation}

For $m=1$, since $E_0=\Omega$ and $e_1=\sigma\xi_0$, $b_1=\mathbb E\|e_1\|^2\le \sigma^2\nu^2$. For $m\ge 1$, the one-step bound~\eqref{eq:proof_one_step} gives
\begin{equation}
b_{m+1}
=
\mathbb{E}\left[\|e_{m+1}\|^2\mathbf 1_{E_m}\right]
\\\le
\rho_r^2\,\mathbb{E}\left[\|e_m\|^2\mathbf 1_{E_m}\right]+\sigma^2\nu^2
\le
\rho_r^2\,b_m+\sigma^2\nu^2,
\end{equation}
where the last inequality uses $E_m\subseteq E_{m-1}$ so that $\mathbb{E}[\|e_m\|^2\mathbf 1_{E_m}]\le b_m$. The bound~\eqref{eq:bm_bound} follows by induction.

Next, $\{\tau_r=m\}=E_{m-1}\cap\{\|e_m\|>r\}$, on which $\|e_m\|^2\mathbf 1_{E_{m-1}}\ge r^2$. Markov's inequality therefore gives
\begin{equation}
\mathbb P(\tau_r=m)
\le
\frac{1}{r^2}\,
\mathbb{E}\left[\|e_m\|^2\mathbf 1_{E_{m-1}}\right]
=
\frac{b_m}{r^2}.
\end{equation}
Summing over $m=1,\dots,M$,
\begin{equation}
    \mathbb P(\tau_r\le M)
    =
    \displaystyle
    \sum_{m=1}^M \mathbb P(\tau_r=m)
    \le
    \displaystyle
    \frac{1}{r^2}\sum_{m=1}^M b_m
    \le
    \frac{M}{r^2}\frac{\sigma^2\nu^2}{1-\rho_r^2},
\end{equation}
which recovers \eqref{eq:tube_exit_bound}. \qed

\begin{remark}[Noise assumptions]
\label{rem:noise_assumptions}
Proposition~\ref{prop:tube_stability} uses the noise only through independence, zero mean and the second-moment bound $\mathbb E\|\xi_m\|^2\le\nu^2$. Any such family qualifies: isotropic Gaussian noise gives $\nu^2=LD$; Laplace or Student-$t$ noise with more than two degrees of freedom, scaled to unit variance per coordinate, gives the same $\nu^2$ and hence the same bound; anisotropic noise with covariance $\Sigma$ enters through $\nu^2=\operatorname{tr}\Sigma$. Families without a second moment, such as Cauchy noise, make the bound vacuous. Appendix~\ref{sec:app_ablation} reports the matching experiment.
\end{remark}

\begin{remark}[Dependence on the inner depth]
\label{rem:uniform_in_M}
The in-tube bound \eqref{eq:tube_pre_exit_bound} is uniform in $M$: the geometric sum is bounded by $\sigma^2\nu^2/(1-\rho_r^2)$ for every $m$, so a deeper inner recursion does not loosen the control on the deviation while the rollout stays in the tube. Only the exit probability \eqref{eq:tube_exit_bound} grows, linearly in $M$, through the union over steps: at fixed $\sigma$ the price of depth is paid in how often the control applies, not in how tight it is.
\end{remark}

\section{Compute-matched selection rules}
\label{app:selection_rules}

The rescoring rules of Table~\ref{tab:main_results} re-compute statistics in a stored particle cloud of $S=16$ unguided ($\beta=0$) trajectories with no further backbone calls. These are: \textit{plurality vote}, the most frequent decoded answer, ties to the lowest index; \textit{terminal-$Q$}, the particle with the largest raw guide logit at step $N$, which is Algorithm~\ref{alg:guided_inference} without resampling and with the tilt applied once at the final step, the best-of-$K$ rule of PTRM \citep{sghaier2026ptrm}; \textit{path-sum tilt}, $\arg\max_s \sum_{n=1}^{N} \log\sigmoid(\tilde q_n^{(s)})$, the Feynman--Kac tilt \eqref{eq:guided_update} without resampling; and \textit{oracle}, which counts a puzzle solved if any particle is correct. Table~\ref{tab:paired_rules} compares the rules on the main-results runs, so the differences are paired; Table~\ref{tab:corruption} degrades the guide with Gaussian noise. The path-sum tilt is the more robust rule: at $\sigma_q=16$ it loses four points on Sudoku DF and under five on PPBench golden, where terminal-$Q$ loses $23$ and $14$. On Maze-Hard, where the guide is flat, neither rule moves by more than two points at any noise level.

\begin{table}[htbp]
  \centering
  \small
  \caption{\textbf{Paired selection rules on the main-results runs} ($5$ seeds $\times$ $5$ folds; exact-solve \%, mean $\pm$ s.d.\ over the $25$ cells; differences are paired within cells and carry $95\%$ confidence intervals). Guided and unguided clouds share seed and fold. Puzzles evaluated: Sudoku-Extreme Test $9{,}984$, DF $1{,}408$; Maze-Hard Test $992$, DF $128$.}
  \label{tab:paired_rules}
  \begin{tabular}{@{}lcccc@{}}
    \toprule
     & Sudoku Test & Sudoku DF & Maze Test & Maze DF \\
    \midrule
    Single rollout ($\beta=0$, first particle) & $84.3\!\pm\!0.7$ & $35.7\!\pm\!2.8$ & $85.2\!\pm\!2.5$ & $13.1\!\pm\!6.5$ \\
    Sequential guiding (ours) & $98.0\!\pm\!0.2$ & $85.9\!\pm\!1.5$ & $85.6\!\pm\!2.8$ & $12.5\!\pm\!5.5$ \\
    Terminal-$Q$, guided cloud & $98.0\!\pm\!0.2$ & $85.9\!\pm\!1.5$ & $86.8\!\pm\!2.4$ & $19.1\!\pm\!9.1$ \\
    Terminal-$Q$, unguided cloud & $98.2\!\pm\!0.3$ & $86.8\!\pm\!1.8$ & $86.8\!\pm\!2.5$ & $18.1\!\pm\!8.3$ \\
    Oracle, guided cloud & $98.0\!\pm\!0.2$ & $86.0\!\pm\!1.5$ & $90.7\!\pm\!2.2$ & $37.1\!\pm\!8.1$ \\
    Oracle, unguided cloud & $98.2\!\pm\!0.3$ & $86.8\!\pm\!1.8$ & $91.1\!\pm\!2.2$ & $37.4\!\pm\!8.7$ \\
    \midrule
    Terminal-$Q$ (unguided) minus ours & $+0.2\!\pm\!0.1$ & $+0.9\!\pm\!0.7$ & $+1.2\!\pm\!0.3$ & $+5.6\!\pm\!2.8$ \\
    Ours minus single rollout & $+13.7\!\pm\!0.3$ & $+50.2\!\pm\!1.0$ & $+0.4\!\pm\!0.6$ & $-0.6\!\pm\!2.6$ \\
    Oracle guided minus oracle unguided & $-0.2\!\pm\!0.1$ & $-0.8\!\pm\!0.7$ & $-0.4\!\pm\!0.4$ & $-0.3\!\pm\!3.1$ \\
    \bottomrule
  \end{tabular}
\end{table}

\begin{table}[htbp]
  \centering
  \footnotesize
  \caption{\textbf{Selection under a corrupted guide.} I.i.d.\ Gaussian noise of s.d.\ $\sigma_q$ is added to every raw guide logit of the stored unguided clouds and the rules are re-read ($8$ draws per cell); exact-solve \%, mean $\pm$ s.d.\ over $5$ folds (Sudoku-Extreme and Maze-Hard, full splits of $10{,}020$, $1{,}427$ and $1{,}000$ puzzles, one seed) or over $5$ seeds (PPBench golden, $42$ puzzles). T is terminal-$Q$, P the path-sum tilt. The online filter was not rerun under corruption.}
  \label{tab:corruption}
  \setlength{\tabcolsep}{3pt}
  \begin{tabular}{@{}ccccccccc@{}}
    \toprule
    & \multicolumn{2}{c}{Sudoku Test} & \multicolumn{2}{c}{Sudoku DF} & \multicolumn{2}{c}{Maze Test} & \multicolumn{2}{c}{PPBench golden} \\
    \cmidrule(lr){2-3}\cmidrule(lr){4-5}\cmidrule(lr){6-7}\cmidrule(lr){8-9}
    $\sigma_q$ & T & P & T & P & T & P & T & P \\
    \midrule
    $0$ & $98.1\!\pm\!0.5$ & $98.1\!\pm\!0.5$ & $88.4\!\pm\!0.6$ & $88.1\!\pm\!0.9$ & $86.9\!\pm\!4.0$ & $85.7\!\pm\!3.8$ & $75.7\!\pm\!2.0$ & $75.7\!\pm\!3.1$ \\
    $4$ & $98.1\!\pm\!0.5$ & $97.9\!\pm\!0.5$ & $88.2\!\pm\!0.7$ & $87.2\!\pm\!1.3$ & $85.8\!\pm\!3.7$ & $86.0\!\pm\!4.3$ & $73.4\!\pm\!1.2$ & $74.4\!\pm\!2.1$ \\
    $8$ & $96.9\!\pm\!0.5$ & $97.8\!\pm\!0.5$ & $81.9\!\pm\!1.0$ & $86.2\!\pm\!1.5$ & $85.8\!\pm\!3.8$ & $85.8\!\pm\!4.0$ & $67.2\!\pm\!0.8$ & $74.2\!\pm\!1.9$ \\
    $16$ & $93.4\!\pm\!0.6$ & $97.4\!\pm\!0.6$ & $65.3\!\pm\!0.9$ & $83.9\!\pm\!1.6$ & $85.3\!\pm\!4.0$ & $85.9\!\pm\!3.7$ & $62.1\!\pm\!1.7$ & $71.2\!\pm\!2.7$ \\
    $32$ & $89.6\!\pm\!0.7$ & $95.8\!\pm\!0.8$ & $51.4\!\pm\!1.1$ & $75.6\!\pm\!1.5$ & $85.7\!\pm\!3.9$ & $85.5\!\pm\!3.9$ & $59.8\!\pm\!1.4$ & $66.9\!\pm\!1.8$ \\
    \bottomrule
  \end{tabular}
\end{table}

\section{PPBench}
\label{app:ppbench}

PPBench \citep{waugh2026ppbench} is a pencil-puzzle benchmark of $300$ golden puzzles across $20$ varieties. PTRM \citep{sghaier2026ptrm} trains a TRM on six of them. We train one with attention on the four whose answers are orientation-free with a single clue layer, Sudoku, Light Up, Nurikabe and Tapa, so that dihedral augmentation needs no token remapping ($33{,}460$ training examples; $H_{\mathrm{cycles}}=3$, $L_{\mathrm{cycles}}=M=4$). Heyawake and Shakashaka are excluded, so our aggregate is not comparable with PTRM's. We evaluate on the $42$ golden puzzles of these types and on a held-out $253$-puzzle validation split from the public pool, disjoint from training and golden. Inference reuses $S=16$, $N=48$, $\sigma=0.3$ and $\beta=0.25$ from the Sudoku-Extreme sweep and fixes $\tau_{\ESS}=0.5$ a priori; nothing is tuned on PPBench. Table~\ref{tab:ppbench_types} gives every rule by type. Over $5$ seeds, sequential guiding lifts validation exact-solve from $62.5\%$ (deterministic) to $77.9\!\pm\!0.6\%$, ahead on every type, and golden from $59.5\%$ to $74.8\!\pm\!2.1\%$. The gain concentrates on Sudoku and Tapa, the types with the most headroom; on golden Sudoku the plurality vote falls below a single rollout while every guided rule roughly doubles it. Guide spread and class gap $\Delta_N$ are large on every type, as on Sudoku-Extreme and unlike Maze-Hard. The deterministic depth sweep is in Table~\ref{tab:depth_det}.

\begin{table}[htbp]
  \centering
  \footnotesize
  \caption{\textbf{PPBench by puzzle type} (exact-solve \%, mean $\pm$ s.d.\ over $5$ seeds; the deterministic TRM is one run). Rescoring rules and the oracle read the stored unguided clouds, and the single rollout here is the mean per-particle rate of those clouds. The last two rows of each block give the guide spread (s.d.\ of the terminal raw logit over particles, averaged over puzzles) and the class gap $\Delta_N$, both in raw-logit units.}
  \label{tab:ppbench_types}
  \setlength{\tabcolsep}{3.5pt}
  \begin{tabular}{@{}lccccc@{}}
    \toprule
    & All & Sudoku & Light Up & Nurikabe & Tapa \\
    \midrule
    \multicolumn{6}{@{}l}{\textit{Validation} ($253$ puzzles: $95$, $63$, $67$, $28$)} \\
    Deterministic TRM & $62.5$ & $47.4$ & $93.7$ & $68.7$ & $28.6$ \\
    Single rollout (mean per particle) & $63.8\!\pm\!0.5$ & $48.9\!\pm\!0.8$ & $89.9\!\pm\!0.4$ & $71.1\!\pm\!1.1$ & $37.9\!\pm\!1.0$ \\
    Plurality vote & $67.7\!\pm\!1.4$ & $51.8\!\pm\!2.2$ & $93.3\!\pm\!0.6$ & $76.1\!\pm\!1.6$ & $43.6\!\pm\!2.7$ \\
    Terminal-$Q$ & $79.5\!\pm\!0.8$ & $79.2\!\pm\!1.5$ & $95.6\!\pm\!1.6$ & $78.2\!\pm\!1.5$ & $47.9\!\pm\!3.6$ \\
    Path-sum tilt & $78.7\!\pm\!0.7$ & $77.5\!\pm\!2.5$ & $95.6\!\pm\!1.6$ & $77.6\!\pm\!1.9$ & $47.9\!\pm\!4.8$ \\
    Sequential guiding (ours) & $77.9\!\pm\!0.6$ & $74.3\!\pm\!1.2$ & $94.6\!\pm\!0.9$ & $78.8\!\pm\!1.9$ & $50.7\!\pm\!3.0$ \\
    Oracle & $82.8\!\pm\!0.7$ & $82.3\!\pm\!1.7$ & $95.9\!\pm\!1.6$ & $83.6\!\pm\!2.1$ & $53.6\!\pm\!2.3$ \\
    Guide spread & $1.770\!\pm\!0.035$ & $2.708\!\pm\!0.089$ & $1.177\!\pm\!0.043$ & $1.123\!\pm\!0.039$ & $1.467\!\pm\!0.060$ \\
    Class gap $\Delta_N$ & $8.2\!\pm\!0.1$ & $9.3\!\pm\!0.1$ & $12.2\!\pm\!0.1$ & $4.2\!\pm\!0.1$ & $5.2\!\pm\!0.1$ \\
    \midrule
    \multicolumn{6}{@{}l}{\textit{Golden} ($42$ puzzles: $15$, $8$, $9$, $10$)} \\
    Deterministic TRM & $59.5$ & $46.7$ & $87.5$ & $88.9$ & $30.0$ \\
    Single rollout (mean per particle) & $57.0\!\pm\!0.8$ & $31.4\!\pm\!3.2$ & $96.6\!\pm\!1.4$ & $81.1\!\pm\!0.5$ & $41.9\!\pm\!2.4$ \\
    Plurality vote & $61.0\!\pm\!3.2$ & $28.0\!\pm\!10.7$ & $100.0\!\pm\!0.0$ & $88.9\!\pm\!0.0$ & $54.0\!\pm\!4.9$ \\
    Terminal-$Q$ & $75.7\!\pm\!1.8$ & $69.3\!\pm\!3.3$ & $100.0\!\pm\!0.0$ & $88.9\!\pm\!0.0$ & $54.0\!\pm\!4.9$ \\
    Path-sum tilt & $75.7\!\pm\!2.8$ & $66.7\!\pm\!4.2$ & $100.0\!\pm\!0.0$ & $88.9\!\pm\!0.0$ & $58.0\!\pm\!7.5$ \\
    Sequential guiding (ours) & $74.8\!\pm\!2.1$ & $62.7\!\pm\!3.7$ & $100.0\!\pm\!0.0$ & $88.9\!\pm\!0.0$ & $60.0\!\pm\!7.1$ \\
    Oracle & $80.5\!\pm\!1.0$ & $69.3\!\pm\!3.3$ & $100.0\!\pm\!0.0$ & $88.9\!\pm\!0.0$ & $74.0\!\pm\!4.9$ \\
    Guide spread & $1.790\!\pm\!0.083$ & $2.631\!\pm\!0.132$ & $1.111\!\pm\!0.262$ & $1.072\!\pm\!0.097$ & $1.719\!\pm\!0.116$ \\
    Class gap $\Delta_N$ & $8.5\!\pm\!0.1$ & $9.8\!\pm\!0.1$ & $11.7\!\pm\!0.6$ & $2.8\!\pm\!0.4$ & $6.2\!\pm\!0.1$ \\
    \bottomrule
  \end{tabular}
\end{table}

\section{Guide quality: calibration and discrimination}
\label{app:guide_quality}

For each benchmark we score every particle of the stored unguided clouds by $\sigmoid(\tilde q)$ at every third outer step and compute, per fold, the ten-bin expected calibration error (ECE) of that score against per-particle exact correctness, and its AUROC as a classifier of correctness, pooling particles over puzzles (Table~\ref{tab:calibration_steps}). Sudoku-Extreme is calibrated and discriminating from step $6$ onwards. Maze-Hard is neither: on Test the AUROC falls from $0.89$ at step $3$ to $0.68$ at step $48$ while the ECE rises to $0.14$, and the terminal mean confidence of $0.995$ sits against a per-particle success rate of $0.856$ (Sudoku-Extreme: $0.961$ against $0.844$); on DF, where few particles are correct, the guide is no better than chance (AUROC $0.46$ to $0.58$). PPBench golden sits between, with AUROC $0.99$ and ECE $0.03$ to $0.08$.

\begin{table}[htbp]
  \centering
  \footnotesize
  \caption{\textbf{Guide calibration and discrimination by outer step} on the stored unguided clouds (mean $\pm$ s.d.\ over $5$ folds, or over $5$ seeds for PPBench golden). ECE: ten-bin expected calibration error of $\sigmoid(\tilde q)$ against per-particle correctness; AUROC: pooled over particles and puzzles. A dash marks a step at which the AUROC is undefined.}
  \label{tab:calibration_steps}
  \setlength{\tabcolsep}{3.5pt}
  \begin{tabular}{@{}llccccc@{}}
    \toprule
    & & $n=3$ & $n=6$ & $n=12$ & $n=24$ & $n=48$ \\
    \midrule
    Sudoku Test & ECE & $0.004\!\pm\!0.001$ & $0.002\!\pm\!0.001$ & $0.001\!\pm\!0.000$ & $0.001\!\pm\!0.000$ & $0.000\!\pm\!0.000$ \\
     & AUROC & $0.994\!\pm\!0.001$ & $1.000\!\pm\!0.000$ & $1.000\!\pm\!0.000$ & $1.000\!\pm\!0.000$ & $1.000\!\pm\!0.000$ \\
    Sudoku DF & ECE & $0.001\!\pm\!0.000$ & $0.001\!\pm\!0.000$ & $0.001\!\pm\!0.000$ & $0.001\!\pm\!0.000$ & $0.001\!\pm\!0.000$ \\
     & AUROC & -- & $1.000\!\pm\!0.000$ & $1.000\!\pm\!0.000$ & $1.000\!\pm\!0.000$ & $1.000\!\pm\!0.000$ \\
    Maze Test & ECE & $0.087\!\pm\!0.027$ & $0.133\!\pm\!0.036$ & $0.138\!\pm\!0.036$ & $0.138\!\pm\!0.036$ & $0.139\!\pm\!0.036$ \\
     & AUROC & $0.894\!\pm\!0.026$ & $0.768\!\pm\!0.044$ & $0.696\!\pm\!0.037$ & $0.697\!\pm\!0.037$ & $0.684\!\pm\!0.033$ \\
    Maze DF & ECE & $0.571\!\pm\!0.030$ & $0.846\!\pm\!0.052$ & $0.864\!\pm\!0.046$ & $0.869\!\pm\!0.045$ & $0.875\!\pm\!0.041$ \\
     & AUROC & $0.485\!\pm\!0.088$ & $0.464\!\pm\!0.127$ & $0.563\!\pm\!0.158$ & $0.563\!\pm\!0.163$ & $0.547\!\pm\!0.181$ \\
    PPBench golden & ECE & $0.063\!\pm\!0.005$ & $0.079\!\pm\!0.004$ & $0.059\!\pm\!0.007$ & $0.042\!\pm\!0.008$ & $0.033\!\pm\!0.002$ \\
     & AUROC & $0.986\!\pm\!0.004$ & $0.993\!\pm\!0.003$ & $0.995\!\pm\!0.002$ & $0.993\!\pm\!0.002$ & $0.989\!\pm\!0.001$ \\
    \bottomrule
  \end{tabular}
\end{table}

\paragraph{Intervening on the guide.}
Alignment in the sense of Lemma~\ref{lem:bce_aligned} depends only on the ranking, which suggests three interventions of increasing cost, none demonstrated here. If calibration fails but ranking survives, a monotone recalibration (temperature or isotonic) repairs the ECE and leaves the lemma intact. If discrimination fails, as on Maze-Hard, no recalibration can help and the guide must be refitted, which is cheap relative to the backbone (Appendix~\ref{sec:app_checkpoint}); this is why \S\ref{sec:discussion} sends Maze-Hard to a refitted guide rather than to recalibration. Where neither suffices, the guide is built bespoke for the task (\S\ref{sec:discussion}).

\section{Depth, particles and cost}
\label{app:depth_cost}

Tables~\ref{tab:depth_det} to~\ref{tab:anytime} collect the depth, particle-count and cost evidence behind \S\ref{sec:experiments}. Deterministic recursion improves slowly with depth: at $N=768$, where its $5{,}376$ backbone calls match the $S=16$, $N=48$ cloud, it reaches $95.7\%$ on a $256$-puzzle Sudoku-Extreme Test subset, still short of the cloud; on PPBench validation it moves from $62.5\%$ to $63.6\%$ and on golden it declines. Sequential guiding passes the full-depth deterministic baseline at step $9$ on Sudoku-Extreme Test, reaches $96.3\!\pm\!0.4\%$ at $N=24$ and is still rising at $N=48$; on Maze-Hard the curve is flat from $N=6$. Over particle count on a $512$-puzzle subset (three folds, hence wide intervals) the guided rate tracks the unguided oracle at every $S$; Figure~\ref{fig:ablation_sensitivity} has the sweep on $2{,}000$ puzzles. Table~\ref{tab:cost} (Appendix~\ref{sec:app_eval_protocol}) gives wall-clock and memory against $S$ on one H100, measured end to end on $64$ puzzles in four batches of $16$ at $N=48$, so it reports batched throughput rather than single-query latency. Finally, halting the stored unguided cloud when $\max_s \sigmoid(\tilde q_n^{(s)})$ first exceeds a threshold, checked every third step, matches the full-horizon terminal-$Q$ rate at a fraction of the budget wherever the guide is calibrated; on Maze-Hard it stops after three steps and gives up around two points.

\begin{table}[htbp]
  \centering
  \small
  \caption{\textbf{Deterministic recursion against outer depth} (exact-solve \%, one run; $N=48$ is the trained horizon and $N=768$ matches the backbone calls of the $S=16$, $N=48$ cloud). Sudoku-Extreme: a $256$-puzzle Test subset; PPBench: validation ($253$ puzzles) and golden ($42$).}
  \label{tab:depth_det}
  \begin{tabular}{@{}lccc@{}}
    \toprule
    $N$ & Sudoku Test (subset) & PPBench validation & PPBench golden \\
    \midrule
    $48$ & $86.7$ & $62.5$ & $59.5$ \\
    $96$ & $90.2$ & $62.5$ & $59.5$ \\
    $192$ & $92.6$ & $62.8$ & $59.5$ \\
    $384$ & $93.8$ & $63.2$ & $57.1$ \\
    $768$ & $95.7$ & $63.6$ & $57.1$ \\
    \bottomrule
  \end{tabular}
\end{table}

\begin{table}[htbp]
  \centering
  \small
  \caption{\textbf{Sequential guiding against outer depth} ($5$ seeds $\times$ $5$ folds; exact-solve \% of the highest-weight particle at step $N$, mean $\pm$ s.d.\ over cells). The deterministic TRM at $N=48$ scores $85.9$, $0.0$, $86.6$ and $0.0$ on the four splits.}
  \label{tab:depth_guided}
  \begin{tabular}{@{}lcccc@{}}
    \toprule
    $N$ & Sudoku Test & Sudoku DF & Maze Test & Maze DF \\
    \midrule
    $6$ & $75.0\!\pm\!0.8$ & $22.0\!\pm\!2.4$ & $85.0\!\pm\!2.3$ & $10.4\!\pm\!5.1$ \\
    $12$ & $91.7\!\pm\!0.6$ & $58.6\!\pm\!3.2$ & $85.2\!\pm\!2.5$ & $11.1\!\pm\!5.5$ \\
    $24$ & $96.3\!\pm\!0.4$ & $77.4\!\pm\!2.8$ & $85.4\!\pm\!2.6$ & $11.3\!\pm\!5.9$ \\
    $36$ & $97.5\!\pm\!0.3$ & $83.1\!\pm\!1.8$ & $85.5\!\pm\!2.7$ & $12.0\!\pm\!5.5$ \\
    $48$ & $98.0\!\pm\!0.2$ & $85.9\!\pm\!1.5$ & $85.6\!\pm\!2.8$ & $12.5\!\pm\!5.5$ \\
    \bottomrule
  \end{tabular}
\end{table}

\begin{table}[htbp]
  \centering
  \small
  \caption{\textbf{Particle count at fixed depth} ($N=48$; a $512$-puzzle Sudoku-Extreme Test subset, $3$ folds of one seed, $\pm$ $95\%$ confidence interval over folds). The deterministic TRM scores $87.3\!\pm\!6.9$ on the same subset, and the $S=1$ guided run is a single stochastic rollout.}
  \label{tab:frontier}
  \begin{tabular}{@{}lccccc@{}}
    \toprule
    & $S=1$ & $S=2$ & $S=4$ & $S=8$ & $S=16$ \\
    \midrule
    Sequential guiding (ours) & $83.5\!\pm\!7.4$ & $90.8\!\pm\!10.1$ & $94.3\!\pm\!7.6$ & $96.0\!\pm\!6.2$ & $98.5\!\pm\!2.9$ \\
    Oracle (unguided cloud) & $86.2\!\pm\!9.1$ & $90.1\!\pm\!9.1$ & $94.7\!\pm\!5.6$ & $97.0\!\pm\!4.1$ & $97.7\!\pm\!3.7$ \\
    \bottomrule
  \end{tabular}
\end{table}

\begin{table}[htbp]
  \centering
  \small
  \caption{\textbf{Anytime halting on the stored unguided clouds.} Halt at the first step $n\in\{3,6,\ldots,48\}$ with $\max_s\sigmoid(\tilde q_n^{(s)})\ge\tau$ and return the particle with the largest score. Exact-solve \% at the stop (mean $\pm$ s.d.\ over $5$ folds, or $5$ seeds for PPBench golden), mean stop step, share of the $N=48$ budget used, and terminal-$Q$ at the full horizon.}
  \label{tab:anytime}
  \begin{tabular}{@{}llcccc@{}}
    \toprule
    & $\tau$ & Exact at stop & Mean stop step & Budget used & Full horizon \\
    \midrule
    Sudoku Test & $0.90$ & $97.4\!\pm\!0.6$ & $6.9$ & $14.4\%$ & $98.1$ \\
     & $0.95$ & $98.0\!\pm\!0.5$ & $7.0$ & $14.5\%$ & \\
     & $0.99$ & $98.1\!\pm\!0.5$ & $7.1$ & $14.7\%$ & \\
    \midrule
    Sudoku DF & $0.90$ & $88.2\!\pm\!0.7$ & $16.8$ & $35.0\%$ & $88.4$ \\
     & $0.95$ & $88.3\!\pm\!0.7$ & $16.8$ & $35.1\%$ & \\
     & $0.99$ & $88.4\!\pm\!0.6$ & $16.9$ & $35.2\%$ & \\
    \midrule
    Maze Test & $0.90$ & $84.6\!\pm\!3.8$ & $3.1$ & $6.4\%$ & $86.9$ \\
     & $0.95$ & $84.8\!\pm\!3.9$ & $3.1$ & $6.4\%$ & \\
     & $0.99$ & $85.3\!\pm\!3.6$ & $3.3$ & $6.9\%$ & \\
    \midrule
    PPBench golden & $0.90$ & $68.6\!\pm\!2.6$ & $12.7$ & $26.4\%$ & $75.7$ \\
     & $0.95$ & $71.9\!\pm\!1.1$ & $13.8$ & $28.8\%$ & \\
     & $0.99$ & $75.2\!\pm\!2.1$ & $16.8$ & $35.0\%$ & \\
    \bottomrule
  \end{tabular}
\end{table}

\end{document}